%% file: root.tex
\documentclass[letterpaper, 10 pt, conference]{ieeeconf}  

\IEEEoverridecommandlockouts                              

\usepackage{amsthm, amsfonts, amsmath}

\usepackage[noend]{algorithmic}
\usepackage[vlined,ruled,linesnumbered]{algorithm2e}

\input{preamble_packages}
\input{preamble_symbols}

\input{shortcuts.tex}

\title{\vspace{-10mm} \LARGE \bf
Online Aggregation of Trajectory Predictors
}

\author{Alex Tong$^1$, Apoorva Sharma$^2$, Sushant Veer$^2$, Marco Pavone$^{2,3}$, Heng Yang$^{1,2}$
\thanks{$^{1}$Harvard University, $^{2}$NVIDIA Research, $^{3}$Stanford University.}%
}

\begin{document}

\maketitle
\thispagestyle{empty}
\pagestyle{empty}



\input{sections/abstract.tex}


\input{sections/introduction.tex}
\input{sections/related_works.tex}
\input{sections/online_aggregation.tex}
\input{sections/experiment.tex}

\input{sections/conclusion.tex}

\bibliographystyle{IEEEtran}  
\bibliography{bibtex/IEEEabrv, bibtex/refs}  

\input{sections/appendix.tex}

\end{document}

%% file: preamble_packages.tex
\usepackage{comment}
\usepackage{siunitx}
\usepackage{relsize}
\usepackage{ifthen}

\usepackage[caption=false]{subfig}

\usepackage[vlined,ruled,linesnumbered]{algorithm2e}
\usepackage{graphics} 
\usepackage{rotating}
\usepackage{color}
\usepackage{enumerate}
\usepackage[T1]{fontenc}
\usepackage{psfrag}
\usepackage{epsfig} 
\usepackage{booktabs}
\usepackage{graphicx,url}
\usepackage{subcaption}
\usepackage{multirow}
\usepackage{array}
\usepackage{latexsym}
\usepackage{amsfonts}
\usepackage{amsmath}
\usepackage{amssymb}
\usepackage{mathtools}
\usepackage{xstring}
\usepackage{multirow}
\usepackage{xcolor}
\usepackage{prettyref}
\usepackage{flexisym}
\usepackage{bigdelim}
\usepackage{breqn} 
\usepackage{listings}
\let\labelindent\relax
\usepackage{enumitem}
\usepackage{xspace}
\usepackage{bm}
\graphicspath{{./figures/}}
\usepackage{tikz}
\usetikzlibrary{matrix,calc}

\usepackage{mdwlist}

\let\itemize\undefined
\makecompactlist{itemize}{stditemize}

%% file: preamble_symbols.tex
\newrefformat{prob}{Problem\,\ref{#1}}
\newrefformat{def}{Definition\,\ref{#1}}
\newrefformat{sec}{Section\,\ref{#1}}
\newrefformat{sub}{Section\,\ref{#1}}
\newrefformat{prop}{Proposition\,\ref{#1}}
\newrefformat{app}{Appendix\,\ref{#1}}
\newrefformat{alg}{Algorithm\,\ref{#1}}
\newrefformat{cor}{Corollary\,\ref{#1}}
\newrefformat{thm}{Theorem\,\ref{#1}}
\newrefformat{lem}{Lemma\,\ref{#1}}
\newrefformat{fig}{Fig.\,\ref{#1}}
\newrefformat{tab}{Table\,\ref{#1}}

\newtheorem{theorem}{Theorem}

\newtheorem{remark}[theorem]{Remark}

\newcommand{\cf}{\emph{cf.}\xspace}

\newcommand{\bdmath}{\begin{dmath}}
\newcommand{\edmath}{\end{dmath}}
\newcommand{\beq}{\begin{equation}}
\newcommand{\eeq}{\end{equation}}
\newcommand{\bdm}{\begin{displaymath}}
\newcommand{\edm}{\end{displaymath}}
\newcommand{\bea}{\begin{eqnarray}}
\newcommand{\eea}{\end{eqnarray}}
\newcommand{\beal}{\beq \begin{array}{ll}}
\newcommand{\eeal}{\end{array} \eeq}
\newcommand{\beas}{\begin{eqnarray*}}
\newcommand{\eeas}{\end{eqnarray*}}
\newcommand{\ba}{\begin{array}}
\newcommand{\ea}{\end{array}}
\newcommand{\bit}{\begin{itemize}}
\newcommand{\eit}{\end{itemize}}
\newcommand{\ben}{\begin{enumerate}}
\newcommand{\een}{\end{enumerate}}

\newcommand{\calT}{{\cal T}}

\newcommand{\eg}{\emph{e.g.,}\xspace}
\newcommand{\ie}{\emph{i.e.,}\xspace}

\newcommand{\hide}[1]{}

\newcommand{\hiddenText}{{\color{gray} hidden text.}}
\newcommand{\hideWithText}[1]{\hiddenText}

\newcommand{\norm}[1]{\left\| #1 \right\|}

\newcommand{\tran}{^{\mathsf{T}}}

\newcommand{\Real}[1]{ { {\mathbb R}^{#1} } }

\newcommand{\blue}[1]{{\color{blue}#1}}

\newcommand{\linkToPdf}[1]{\href{#1}{\blue{(pdf)}}}
\newcommand{\linkToPpt}[1]{\href{#1}{\blue{(ppt)}}}
\newcommand{\linkToCode}[1]{\href{#1}{\blue{(code)}}}
\newcommand{\linkToWeb}[1]{\href{#1}{\blue{(web)}}}
\newcommand{\linkToVideo}[1]{\href{#1}{\blue{(video)}}}
\newcommand{\linkToMedia}[1]{\href{#1}{\blue{(media)}}}
\newcommand{\award}[1]{\xspace} 



%% file: shortcuts.tex
\renewcommand{\norm}[1]{\left\lVert #1 \right\rVert}

\newcommand{\cbrace}[1]{\left\{#1\right\}}
\newcommand{\sym}[1]{\mathbb{S}^{#1}}

\newcommand{\bmat}{\left[ \begin{array}}
\newcommand{\emat}{\end{array}\right]}

\newcommand{\pd}[1]{\sym{#1}_{++}}
\newcommand{\parentheses}[1]{\left(#1\right)}

\newcommand{\nuscenes}{\textsc{nuScenes}\xspace}
\newcommand{\lyft}{\textsc{Lyft}\xspace}
\newcommand{\regret}{\mathrm{Regret}}
\newcommand{\squint}{\textsc{squint}\xspace}

%% file: sections/abstract.tex

\begin{abstract}
    Trajectory prediction, the task of forecasting future agent behavior from past data, is central to safe and efficient autonomous driving. A diverse set of methods (\eg rule-based or learned with different architectures and datasets) have been proposed, yet it is often the case that the performance of these methods is sensitive to the deployment environment (\eg how well the design rules model the environment, or how accurately the test data match the training data). Building upon the principled theory of \emph{online convex optimization} but also going beyond convexity and stationarity, we present a lightweight and model-agnostic method to aggregate different trajectory predictors online. We propose treating each individual trajectory predictor as an ``expert'' and maintaining a probability vector to mix the outputs of different experts. Then, the key technical approach lies in leveraging online data --the true agent behavior to be revealed at the next timestep-- to form a convex-or-nonconvex, stationary-or-dynamic loss function whose gradient steers the probability vector towards choosing the best mixture of experts. We instantiate this method to aggregate trajectory predictors trained on different cities in the \nuscenes dataset and show that it performs just as well, if not better than, any singular model, even when deployed on the out-of-distribution \lyft dataset.

\end{abstract}

%% file: sections/introduction.tex
\section{Introduction}
\label{sec:intro}

Predicting future trajectories of traffic agents is a central task in autonomous driving. Traditionally, rule-based methods were used for their simplicity, efficiency, and explainability~\cite{censi19icra-liability,veer23icra-receding,helou21iros-reasonable}. In contrast, recent studies have shown that deep learning-based approaches~\cite{salzmann20eccv-trajectron++,chen22cvpr-scept,yuan21iccv-agentformer} achieve superior performance due to their ability to learn spatial relationships and inter-agent dynamics from rich high-dimensional data.

However, each trajectory predictor is often trained on a specific dataset with a specific architecture, raising challenges when encountering new unseen datasets, as distribution shifts can significantly degrade a model's test-time performance. A standard approach to mitigate this issue is \emph{data augmentation}, \ie combine all possible datasets and train a large model that hopefully becomes a generalist. While the promise of general foundational models trained from large datasets has (mostly) been fulfilled by large language and vision models~\cite{bommasani21arxiv-opportunities}, their out-of-distribution (OOD) generalization ability remains to be rigorously evaluated~\cite{wang2023decodingtrust,nguyen2023out,liu2024survey}. Moreover, in the context of trajectory prediction, it is unclear yet whether such a \emph{globally} dominant model exists, as the distribution of future agent behavior can be heavily influenced by local traffic rules and driving norms across geographical locations.

In this paper, we explore the use of \emph{online aggregation} of trajectory predictors as an alternative approach to out-of-distribution generalization. In statistical learning terms, our framework can be considered an example of \emph{boosting}~\cite{schapire99ijcai-brief,meir03ml-introduction}.  Particularly, we treat each trajectory predictor as a black-box ``expert'' --whether the predictor was hand-crafted or learned, what architecture was used for learning, and which dataset it was trained on are all unknown to us-- and use a probability vector to mix the outputs of different experts. The technical problem then becomes how to update the probability vector such that the \emph{mixture-of-experts} (MoE) model remains robust to distribution shifts.

To solve this problem, we leverage theory from the emerging field of \emph{online convex optimization} (OCO)~\cite{orabona19book-modern}. In particular, OCO considers a dynamic game between a player and an adversarial environment. In each round $t$, the player outputs a decision $z_t$, the environment chooses a (potentially adversarial) convex loss function $l_t(z_t)$ and reveals the gradient $g_t = \nabla l_t(z_t)$. The goal of the player is to design an algorithm that takes in $g_t$ and outputs $z_t$ to minimize the notion of \emph{regret}:
\bea \label{eq:regret}
\regret_T(l_{1:T},u) := \sum_{t=1}^T l_t(z_t) - \sum_{t=1}^T l_t(u),
\eea 
where $T$ is the horizon of the game and $u$ is any fixed comparator. Intuitively, we wish the regret to be \emph{sublinear} in $T$, then the algorithm will converge to the best fixed decision in hindsight. Classical and recent literature in online learning established that \emph{gradient descent} and its variants can indeed attain sublinear regret, see~\cite{orabona19book-modern,zhang2024improving} and references therein.

{\bf Contribution}. 
We borrow the OCO framework to design a principled and lightweight framework for online aggregation of multiple trajectory predictors. We first assume the output of each trajectory predictor is a Gaussian mixture model (GMM). Then, using the OCO language presented above, the player's decision $z_t$ is the probability vector, resulting in an MoE model that is a mixture of GMMs, which is a new GMM. The loss function $l_t$ uses the true agent state revealed at time $t$ to evaluate the performance of the MoE model. The player finally uses the gradient of $l_t$ to perform ``online gradient descent'' to update the probability vector. When the loss function is chosen as the negative probability loss, it becomes \emph{linear} in $z_t$. In this case, the most well-known algorithm for online minimization of a linear function over $z_t$ (a probability vector) is called \emph{exponentiated gradient} (EG)~\cite{orabona19book-modern}. However, we found the empirical convergence rate of EG is too slow to be practical. To mitigate this issue, we depart from EG and choose the recent \squint algorithm~\cite{koolen2015secondorder} with Cutkosky clipping~\cite{cutkosky2019artificial} to update the probability vector online, which demonstrated roughly 25 times faster empirical convergence (see Appendix~\ref{app:eg-squint}).

We then go beyond the convexity and stationarity assumptions in OCO and engineer two heuristics for a nonconvex loss function and nonstationary environments (\ie the distribution shift happens more than once). First, when the loss function is chosen as the top-$k$ minimum first displacement error (minFRDE$_{k}$, a variant of a common performance metric in the literature), we approximate it using a differentiable surrogate~\cite{grover2019stochastic} and then apply \squint. Second, if the environment is nonstationary, we apply a discount factor when computing losses from previous steps~\cite{zhang2024discountedn}, which allows the algorithm to gracefully forget the past and adapt to new distribution shifts. 

To study the performance of our algorithm, we pretrain three trajectory predictors on the Boston, Singapore, and Las Vegas splits of the \nuscenes dataset~\cite{caesar2020nuscenes}. Together with a rule-based trajectory predictor~\cite{veer23arxiv-mpf}, we perform online aggregation on two out-of-distribution datasets: (i) the Pittsburgh split of \nuscenes and (ii) the \lyft dataset~\cite{houston2021one}. We show that the MoE model performs on par or better than any singular model on various performance metrics. 

To summarize, our contributions are:
\begin{itemize}
    \item We introduce a framework for online aggregation of multiple trajectory predictors to gain robustness against distribution shifts. Notably, our framework is agnostic to the design of individual predictors, as long as each of them outputs a distribution of trajectories.

    \item We apply the theory of online convex optimization to design a lightweight algorithm for online aggregation. Specifically, we identified that the classical EG algorithm in OCO is impractical for online aggregation in trajectory prediction, whereas the recent \squint algorithm enables fast adaptation.
    
    \item We extend the OCO framework to handle nonconvexity and nonstationarity, allowing the online aggregation algorithm to optimize for diverse scenarios.
\end{itemize}

%% file: sections/related_works.tex
\section{Related Works}
\label{sec:related}

{\bf Trajectory Prediction}. Early works on trajectory prediction focused on rule-based models, operating under the assumption that humans typically conform to a set of rules and customs \cite{6696982}. 
However, the performance of rule-based predictors relies heavily on the completeness and soundness of the predefined rules. Furthermore, the ``rigidity'' of the rules can fail to capture the nuanced and stochastic nature of human driving, leading to an inability to reason about uncertainty and adapt to multimodal outcomes in complex interactions. 
Learning-based methods can address these limitations by leveraging data to train generative models~\cite{salzmann20eccv-trajectron++,chen22cvpr-scept,yuan21iccv-agentformer} that reason about uncertainty and provide multimodal outcomes. 
Various architectures have been explored, \eg recurrent neural networks (RNNs)~\cite{7780479, kamenev2022predictionnet}, transformers~\cite{yuan21iccv-agentformer}, variational auto-encoders (VAEs) \cite{salzmann20eccv-trajectron++}, generative adversarial networks (GANs) \cite{gupta2018social}, and even large language models (LLMs) \cite{bae2024can}.  

{\bf Trajectory Predictor Fusion}. 
Despite their ability to model uncertainty and multimodal behaviors, learned predictors often struggle to generalize to OOD scenarios, where the test distribution differs from the training data.

This has led to recent works combining rule-based and learned predictors in series~\cite{song2021learning, li2023planninginspired, pmlr-v155-li21b} and in parallel~\cite{veer23arxiv-mpf, sun2021complementing, patrikar2024rulefuser}. While our proposed architecture is also in parallel, it does not use a Bayesian method for fusion and instead builds upon online convex optimization.

{\bf Online Learning in Robotics}.
A notable application of online convex optimization (OCO) in robotics is \emph{online control}, which dates back to at least~\cite{agarwal19icml-online}. This approach reformulates online optimal control as an OCO problem, leveraging algorithmic insights from the OCO literature~\cite{hazan22-introduction}. Building on these ideas, adaptive online learning has enabled parameter-free optimizers for lifelong reinforcement learning~\cite{muppidi2024pick}. OCO has also proven valuable for \emph{online adaptive conformal prediction}~\cite{gibbs21-adaptive}, where it helps construct prediction sets that guarantee coverage despite distribution shifts~\cite{zhang2024discountedn,bhatnagar2023improved}. Inspired by these successes, we investigate OCO's application to aggregating trajectory predictors.

%% file: sections/online_aggregation.tex
\section{Online Aggregation}
\label{sec:online-agg}

In this section, we present our framework for online aggregation of trajectory predictors. We first define the problem setup (\S\ref{sec:problem-setup}) and spend most of the content on presenting the algorithm for the case of convex loss and stationary environment (\S\ref{sec:convex-stationary}). After that, the extension to nonconvex loss and nonstationary environment becomes natural (\S\ref{sec:nonconvex-loss} and \S\ref{sec:nonstationary}). Fig.~\ref{fig:diagram} overviews our approach.
\vspace{-2mm}
\input{sections/fig-diagram}
\vspace{-6mm}

\subsection{Problem Setup}
\label{sec:problem-setup}
We are given $N$ trajectory predictors treated as black-box experts.
We assume each trajectory predictor outputs a multimodal distribution parameterized as a categorical normal prediction, \ie it predicts a probability vector $p \in \Delta_L$\footnote{$\Delta_n := \{ v \in \Real{n} : v_i \geq 0, i=1,\dots,n, \sum_{i=1}^n v_i =1 \}$ denotes the $n$-D probability simplex.} over $L$ possible modes of future behavior. The $j$-th mode is modeled as a Gaussian distribution with mean $\mu_j \in \mathbb{R}^{K\times3}$ (denoting the $xy$ position and heading $\theta$ of the agent in the next $K$ timesteps) and precision $h_j \in \mathbb{R}^{K\times3}$ (\eg $h_j(k,:) = [1/\sigma_x^2, 1/\sigma_y^2,1/\sigma_{\theta}^2]^T$ represents the reciprocal of the variances in $x,y,\theta$ coordinates at the $k$-th prediction step). 
Let $\Sigma_j$ be the covariance matrix corresponding to $h_j$,
we can write the output of the $i$-th predictor as a Gaussian Mixture Model (GMM):
\bea \label{eq:individual-GMM}
\cbrace{ (p_1^i, \mu_1^i, \Sigma_1^i),\cdots,(p_L^i, \mu_L^i, \Sigma_L^i)  }, \quad i=1,\dots,N
\eea
where $p^i_j \in \Delta_L, \mu^i_j \in \Real{K \times 3}, \Sigma_j^i \in (\pd{3})^K, j=1,\dots,L$ ($\pd{3}$ denotes the set of $3\times 3$ positive definite matrices). Note that we use the superscript $i$ to index the $i$-th expert (trajectory predictor) and the subscript $j$ to index the $j$-th mode in the GMM.

\begin{remark}[Generality]\label{remark:general}
    In principle, our framework only requires each expert to output a distribution of trajectories that can be sampled from, and the distribution needs not be a GMM, see Appendix~\ref{app:generality}. Here we choose to present our framework by assuming each expert outputs a GMM for two reasons. (i) Many recent state-of-the-art trajectory predictors use GMMs to represent multimodal distributions, such as Trajectron++~\cite{salzmann20eccv-trajectron++}, Wayformer~\cite{nayakanti2023wayformer}, and MTR++~\cite{shi2024mtr++}. (ii) GMMs are easier to interpret than arbitrary distributions, and do not require sampling for online aggregation.
\end{remark}

{\bf The Mixture-of-Experts Model}.
Given $N$ predictions of the form~\eqref{eq:individual-GMM}, we aggregate them by forming a mixture-of-experts (MoE) model. Particularly, the MoE model maintains a probability vector $\alpha \in \Delta_N$ where each $\alpha_i$ is used to weigh the importance of the $i$-th expert.  Given $\alpha$, the MoE model outputs a new GMM model that is the mixture of $N \times L$ Gaussians
\bea \label{eq:moe}
\Gamma(\alpha) := \cbrace{ (\alpha_i p^i_j, \mu^i_j, \Sigma^i_j) }_{i=1,j=1}^{i=N,j=L}.
\eea
The goal of online aggregation is then to use online information to dynamically update $\alpha$. 

\subsection{Convex Loss and Stationary Environment}
\label{sec:convex-stationary}
We use OCO to update $\alpha$ in the MoE model~\eqref{eq:moe}. Let $\alpha_t$ be the probability vector ``played'' by the MoE model at time $t$, and $x_t \in \Real{3}$ be the true agent state revealed to the MoE model before it plays the next probability vector. 
A first choice for updating $\alpha_t$ given $x_t$ is to seek an $\alpha$ that \emph{maximizes} the probability of $x_t$ given $\Gamma(\alpha_t)$:
\bea\label{eq:prob-x-given-Gamma}
p(x_t | \Gamma(\alpha_t)) = \sum_{i,j} \alpha_{i,t} p_{j,t}^i g(x_t | \mu^i_{j,t}, \Sigma^i_{j,t})
\eea
where $g(x_t | \mu^i_{j,t}, \Sigma^i_{j,t})$ denotes the probability density function (PDF) of a Gaussian distribution.
In \eqref{eq:prob-x-given-Gamma}, we used the fact that the PDF of a GMM can be written as the weighted sum of the PDFs of its individual Gaussian components. Note that in~\eqref{eq:prob-x-given-Gamma}
we added a subscript ``$t$'' to $(p,\mu,\Sigma)$ to make explicit their dependence on the time step. 
Finding the weights $\alpha$ that maximizes the probability $p(x_t | \Gamma(\alpha_t))$ is equivalent to finding the weights $\alpha$ that minimizes $-p(x_t | \Gamma(\alpha_t))$:
\bea \label{eq:probability-loss}
\ell_t(\alpha_t) = - p(x_t | \Gamma(\alpha_t)).
\eea 

Clearly, the loss function is linear in $\alpha_t$ and its gradient with respect to $\alpha_t$ is 
\bea\label{eq:gradient-prob}
\nabla_{\alpha_{t,i}} \ell_t(\alpha_t) = - \sum_{j} p^{i}_{j,t} g(x_t | \mu_{j,t}^i, \Sigma^i_{j,t}), \ i=1,\dots,N.
\eea

{\bf The \squint Algorithm}. The de-facto algorithm for online minimization of the linear function $\ell_t(\alpha_t)$ in~\eqref{eq:probability-loss} subject to $\alpha_t$ being a probability vector is the \emph{exponentiated gradient} (EG) algorithm~\cite{orabona19book-modern}. However, applying EG in the context of aggregating trajectory predictors is impractical because EG exhibits slow convergence (see Appendix~\ref{app:eg-squint} for a description of the EG algorithm and results showing its slow convergence). To mitigate this issue,
we introduce \squint~\cite{koolen2015secondorder} as the algorithm to update $\alpha$, whose pseudocode is given in Algorithm~\ref{alg:squint} and follows the official implementation.\footnote{\url{https://blog.wouterkoolen.info/Squint_implementation/post.html}}  Without going too deep into the theory, we provide some high-level insights.

\begin{itemize}
\item  {\bf Goal of learning}. Since the loss function is linear, the OCO problem reduces to \emph{hedging}~\cite{freund1997decision}, or \emph{learning with expert advice}~\cite{orabona19book-modern}. In this setup, the $i$-th entry of the gradient ${g_t} \in \Real{N}$ in line~\ref{line:clip-gradient} (for the discussion here let us first assume $g_t = \tilde{g}_t$) represents the loss of the $i$-th expert, for $i=1,\dots,N$. For online aggregation, one can check that the gradient in~\eqref{eq:gradient-prob} represents the probability of $x_t$ under the $i$-th GMM and measures how well the $i$-th expert predicts the ground truth. By playing a MoE model using the probability vector $\alpha_t$, we suffer a weighted sum of individual expert losses, namely $\alpha_t\tran g_t$. The \emph{instantaneous regret} $r_t$ defined in line~\ref{line:ins-regret} measures the \emph{relative performance} of the MoE model with respect to each expert at time $t$, and the cumulative regret $R_t = \sum_{\tau=1}^t r_t$ computed in line~\ref{line:updateRV} measures the relative performance over the time period until $T$. The goal of the learner (\ie the MoE model) is to minimize each entry of $R_t$, \ie to ensure that the regret with respect to \emph{any individual expert} is small. By doing so, the learner is guaranteed to perform \emph{as well as the best expert} (without knowing which expert is the best \emph{a priori}).

\item {\bf Bayes update}. To achieve the goal of small regret, the \squint algorithm essentially performs Bayes update. To see this, note that the algorithm starts with a prior distribution of expert weights $\alpha_1$, which represents the learner's belief of which expert is the best, before receiving any online information. After the game starts, \squint maintains the regret vector $R_t$, together with the \emph{cumulative regret squared} vector $V_t = \sum_{\tau=1}^t r_\tau \circ r_\tau$ computed in line~\ref{line:updateRV}. The main update rule is written in~\eqref{eq:squint-update-alpha} where the function $\xi(\cdot,\cdot)$ is a potential function called \squint evidence. Note that the crucial difference between \squint and the vanilla EG algorithm (see Appendix~\ref{app:eg-squint}) is that \squint uses second-order information, \ie $V_t$. With this interpretation, we recognize~\eqref{eq:squint-update-alpha} as an instance of Bayes' rule with prior $\alpha_1$ and evidence measured by $\xi$. The detailed proof of why~\eqref{eq:squint-update-alpha} guarantees small regret is presented in the original paper~\cite{koolen2015secondorder} and beyond the scope here due to its mathematical involvement. 

\item {\bf Gradient clipping} \ \ An assumption of the hedging setup is that individual expert losses need to be bounded, which may not hold for our problem. Therefore, in lines~\ref{line:Gt}-\ref{line:clip-gradient}, we employ the gradient clipping method proposed in~\cite{cutkosky2019artificial} to rescale the gradients. The gradient clipping technique has been a popular method in training deep neural networks~\cite{pascanu2012understanding,gorbunov2020stochastic,zhang2020adaptive}.

\end{itemize}

\input{sections/alg-squint}

The \squint algorithm guarantees sublinear regret~\cite[
Theorem 4]{koolen2015secondorder}. The loss function $\ell_t(\cdot)$ is linear in $\alpha$ and $\alpha$ lives in the probability simplex, which implies the optimal $\alpha$ in hindsight must be a one-hot vector that assigns $1$ to the best expert and $0$ to other experts~\cite{convexanalysis}. Therefore, we conclude that \squint guarantees convergence to the best expert in hindsight, as we will empirically show in \S\ref{sec:experiment}.

\subsection{Nonconvex Loss}
\label{sec:nonconvex-loss}

The probability loss~\eqref{eq:probability-loss} intuitively considers the ``average'' performance of the MoE model in the sense that all the Gaussian modes are included (penalized) in computing the loss. In practice, however, we may not care about all modes of the MoE model but rather its top-$k$ modes. Formally, let $\alpha_t$ be the decision of the MoE model at time $t$ and denote $\phi_t=(\phi_{1,t},\dots,\phi_{k,t}) \in [NL]$ as the indices of the top-$k$ modes in the MoE model~\eqref{eq:moe}
\bea \label{eq:argtopk}
\phi_t=(\phi_{1,t},\dots,\phi_{k,t}) = \mathrm{argtopk}\{\alpha_{i,t} p^i_{j,t} \}_{i=1,j=1}^{i=N,j=L},
\eea
\ie we rank the weights of all $NL$ modes --each mode has weight $\alpha_{i,t} p^i_{j,t}$-- and then choose the indices of the largest $k$ modes. It is worth noting that $\phi_t$ is a function of $\alpha_t$. Let $x_t$ be the true agent state revealed at time $t$, the minimum first displacement error (minFRDE$_k$) is computed as 
\bea \label{eq:minFRDEk}
\ell_t(\alpha_t) = \min_{i=1,\dots,k} \cbrace{ \norm{x_t - \mu(\phi_i)} }, 
\eea 
where $\mu(\phi_i)$ denotes the mean of the Gaussian distribution with index $\phi_i$.

The minFRDE$_k$ loss in~\eqref{eq:minFRDEk} is, unfortunately, not differentiable in $\alpha_t$ due to the non-differentiability of the ``$\mathrm{argtopk}$'' operation in~\eqref{eq:argtopk}. To fix this issue, we use the $\mathrm{softsort}$ surrogate introduced in~\cite{prillo2020softsort}, and additionally, employ a smooth approximation of the ``$\min$'' operator, \ie
$
\mathrm{softmin}(x) = -\frac{1}{\beta} \log\left(\sum_{i=0}^k \exp(-\beta x_i)\right) 
$
so that the gradients remain non-zero for all $\alpha_{i,t}$ that do not contain the top-$k$ modes ($\beta>0$ is known as the temperature scaling factor). 

In summary, for the nonconvex and nonsmooth minFRDE$_k$ loss~\eqref{eq:minFRDEk}, we perform a smooth approximation using $\mathrm{softsort}$ and $\mathrm{softmin}$, then we directly apply \squint in Algorithm~\ref{alg:squint}.

\subsection{Nonstationary Environment}
\label{sec:nonstationary}

The OCO framework presented so far (\ie the regret in~\eqref{eq:regret} and the \squint algorithm) implicitly assumes a \emph{static} environment. To see this, note that minimizing the regret~\eqref{eq:regret} makes sense if and only if a good comparator $u$ exists in hindsight, which is typically the assumption of a stationary environment. However, this may not be the case for trajectory prediction. Imagine a vehicle on a road trip from Boston to Pittsburgh, or a driver who usually drives in suburban areas but occasionally goes to the city: in both situations the environments are nonstationary. 

Fortunately, there exists an easy fix, as pointed out by~\cite{zhang2024discountedn}. The intuition there is simple: in nonstationary online learning, the key is to gracefully forget about the past and focus on data in a recent time window. Technically, \cite{zhang2024discountedn} shows any stationary online learning algorithm can be converted to a nonstationary algorithm by choosing a discount factor when computing the loss functions. We borrow this intuition and design a nonstationary algorithm by simply adding a discount factor $\lambda < 1$. In particular, the discounted \squint algorithm replaces line~\ref{line:updateRV} of Algorithm~\ref{alg:squint} with
\bea \label{eq:discounted-squint}
R_{t+1} = \lambda R_t + r_t, \quad V_{t+1} = \lambda^2 V_t + r_t \circ r_t.
\eea 

%% file: sections/fig-diagram.tex

\begin{figure}[h]
    \begin{minipage}{\textwidth}
        \includegraphics[width=0.475\columnwidth]{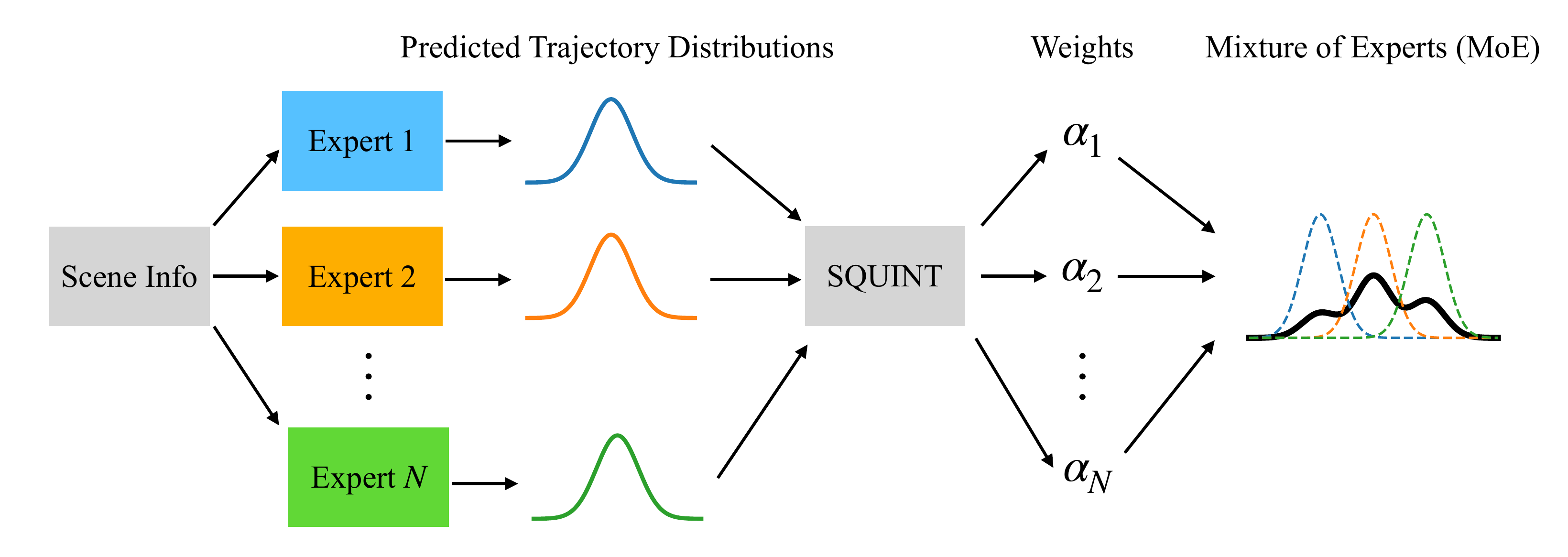}
    \end{minipage}
     \caption{Online aggregation of trajectory predictors.}
    \label{fig:diagram}
\end{figure}

%% file: sections/alg-squint.tex
\setlength{\textfloatsep}{0pt}
\begin{algorithm}[hbt!]
\SetKwInput{KwInput}{Input}
\SetKwInput{KwInit}{Initialize}
\caption{\squint with Cutkosky Clipping}\label{alg:squint}
\KwInput{A prior distribution of the weights $\alpha_1$ for each expert}
\KwInit{Cumulative regret $R_1=0 \in \Real{N}$, Cumulative regret squared $V_1=0 \in \Real{N}$}
\For{$t=1,2,\dots, T$}{
    Compute $\ell_t$ as in \eqref{eq:probability-loss} and define the gradient $\tilde{g}_t = \nabla_{\alpha_t}(\ell_t) \in \Real{N}$ 
    
    Set $G_t = \max (\max_{i \in [N]}(|\tilde{g}_{t,i}|), G_{t-1})$ ($\tilde{g}_{t,i}$ denotes the $i$-th element of $\tilde{g}_t$) \label{line:Gt}
    
    
    Define the clipped gradient $g_t = \frac{(\tilde{g}_t / G_t) + 1}{2} \in \Real{N}$ (element-wise division) \label{line:clip-gradient}
    
    Define the instantaneous regret $r_t = \alpha_t\tran g_t - g_t \in \Real{N}$\label{line:ins-regret}
    
    Update 
    \bea\label{eq:squint-update-alpha}
    \alpha_{t+1,i} = \frac{\alpha_{1,i} \xi(R_{t,i}, V_{t,i})}{\sum_i \alpha_{1,i} \xi(R_{t,i}, V_{t,i})}, \quad i=1,\dots,N
    \eea
    where
    \bea\label{eq:squint-error-function}
    \hspace{-4mm} \xi(R,V) = \sqrt{\pi} \exp{\left( \frac{R^2}{4V}\right)} \frac{\mathrm{erfc}(\frac{-R}{2\sqrt{V}})-\mathrm{erfc}(\frac{V-R}{2\sqrt{V}})}{2\sqrt{V}}
    \eea
    and $\mathrm{erfc}$ is the complementary error function $\mathrm{erfc}(x) := 1 - \frac{2}{\sqrt{\pi}} \int_0^x e^{-t^2} dt$
    
    Update $R_{t+1} = R_{t} + r_t$ and $V_{t+1} = V_{t} + r_t \circ r_t$ (``$\circ$'' denotes element-wise multiplication)\label{line:updateRV}
}
\end{algorithm}

%% file: sections/experiment.tex
\section{Experiments}
\label{sec:experiment}

We study the performance of our framework on the \nuscenes dataset~\cite{caesar2020nuscenes} and the \lyft dataset ~\cite{houston2021one}.

Since trajectory prediction is a well-studied subfield of autonomous driving, our goal here is not to study whether our framework can deliver state-of-the-art performance against other methods, but rather to study the dynamics of our algorithm in the context of trajectory prediction, \ie to investigate whether the OCO machinery and its variants can be a useful tool for trajectory prediction.

{\bf Pretraining Trajectory Predictors on City Splits}. 
Our set of expert trajectory predictors contains three trajectory prediction models that share a common transformer-based architecture, but are trained on three location-based subsets of the training split of the NuPlan dataset \cite{caesar2022nuplan}: Boston, Singapore, and Las Vegas, with the Pittsburgh subset reserved for evaluation. The model architecture uses the factored spatio-temporal attention mechanism used in the Scene Transformer \cite{ngiam2021scene} to encode the agent and neighbor histories, as well as a set of fixed anchor future trajectories. The models predict a mixture model, which consists of a categorical distribution over Gaussian modes whose means are computed as corrections to the fixed anchor trajectories.

{\bf Rule-based Predictor}.
The rule hierarchy (RH) predictor~\cite[Section IV.A]{patrikar2024rulefuser} serves as our rule-based predictor, encoding expected agent behaviors through a prioritized set of rules~\cite{veer23icra-receding}. These rules are arranged in decreasing order of importance: \texttt{collision avoidance}, \texttt{stay within drivable area}, \texttt{traffic lights}, \texttt{speed limit}, \texttt{minimum forward progression}, \texttt{stay near the current lane polyline}, \texttt{stay aligned with the current lane polyline}. The hierarchy is accompanied by a reward function $W$ that quantifies how well a trajectory adheres to these rules.
Trajectory prediction with the RH predictor involves the following steps: First, we generate a trajectory tree that represents potential maneuvers that the vehicle can execute. Then, using the rule hierarchy reward $W$, we compute the reward for each branch of the trajectory tree. Finally, we treat these rewards as negative energy of a Boltzmann distribution to generate a discrete probability distribution over the branches of the trajectory tree. Sampling trajectories from this discrete probability distribution gives us the predicted trajectories. 
Interested readers can find more details in \cite{patrikar2024rulefuser}.

{\bf Performance Metrics}. 
We evaluate the model's performance on three popular performance metrics.

\begin{enumerate}
    \item Top-$k$ Minimum Average Displacement Error (minADE$_k$). Using the same notation from (minFRDE$_k$), we denote $\phi_t=(\phi_{1,t},\dots,\phi_{k,t}) \in [NL]$ as the indices of the top-$k$ modes in the MoE model. 
The minimum average displacement error (minADE$_k$) is then computed as
\begin{align*}
    \min_{i = 1, \dots, k} \cbrace{ \frac{1}{K} \sum_{\tau=1}^{K} \norm{x_t(\tau)  - \mu(\phi_i, \tau)} }, 
\end{align*} where $K$ is the prediction horizon.

    \item Top-$k$ Minimum Final Displacement Error (minFDE$_k$). Calculating minFDE$_k$ is the same as calculating minADE$_k$, but only considering the error of the final predicted timestep:
    \begin{align*}
        \min_{i = 1, \dots, k} \cbrace{  \norm{x_t(K)  - \mu(\phi_i, K)} }. 
    \end{align*}

    \item Negative Log Likelihood (NLL). The negative log likelihood of observing the groundtruth $x_t$ under the predicted (Mixture of Gaussian) distribution, \ie applying $-\log(\cdot)$ to \eqref{eq:prob-x-given-Gamma}.

\end{enumerate}

These performance metrics are averaged over a sliding window of length $500$ (stationary) and $5000$ (nonstationary) for better visualization.

\subsection{Stationary Environment}

\input{sections/fig-stationary}

\input{sections/fig-lyft}

{\bf Convex Loss}. 
We first evaluate the performance of MoE using the convex probability loss~\eqref{eq:probability-loss}. Since the rule-based trajectory predictor does not output covariance in its predictions, we only aggregate the three learned predictors. To simulate distribution shifts, we tested the MoE on the Pittsburgh environment of \nuscenes. Fig.~\ref{fig:stationary}(a) plots the results. We have the following observations. (i) Looking at Fig.~\ref{fig:stationary}(a) bottom, as the theory predicted, the \squint algorithm converges to choosing the best expert in hindsight, which is the model pretrained in the Boston split (this intuitively makes sense because Boston is similar to Pittsburgh). (ii) Looking at Fig.~\ref{fig:stationary}(a) top and middle, we see the MoE's performance, in terms of both NLL and minADE, is better than any of the singular models, and is even close to the oracle Pittsburgh model (that is trained on Pittsburgh). Similar observations hold for the minFDE performance and are shown in Appendix~\ref{app:experiments}. 
We then aggregate the pretrained models and test on the \lyft dataset~\cite{houston2021one}, whose results are shown in Fig.~\ref{fig:lyft}(a). We observe that the MoE model quickly converges to choosing the Las Vegas model as the best expert, and the performance of the MoE model closely tracks that of the best expert Las Vegas.

{\bf Nonconvex Loss}.
We then evaluate the MoE's performance using the nonconvex minFRDE$_k$ loss. In this case, we have two MoEs, one that only aggregates the learned predictors (denoted MoE w/o RB), and the other that aggregates the rule-based predictor in addition (denoted MoE with RB). Fig.~\ref{fig:stationary}(b) plots the results of aggregation on the Pittsburgh dataset. We have the following observations. (i) Looking at Fig.~\ref{fig:stationary}(b) bottom, we see that, unlike the case of convex loss, the probability vector $\alpha$ does not converge. (ii) Even though the probability vector does not converge, the MoE's performance is still better than all the other singular models when evaluated using the minADE performance metric (\cf Fig.~\ref{fig:stationary}(b) middle). (iii) When evaluated using the NLL performance metric (in which case we only evaluate MoE w/o RB because the rule-based predictor does not output covariance), the MoE's performance is better than the Las Vegas model and close to the Boston and Singapore model, though slightly worse. This is expected because the minFRDE loss function does not encourage the minimization of the NLL metric. Fig.~\ref{fig:lyft}(b) plots the results of aggregation on the \lyft dataset. We observe that since the \lyft dataset is very different from \nuscenes, pretrained models on \nuscenes perform poorly but the rule-based predictor maintains good performance. In this case, the MoE model quickly converges to choosing the rule-based model as the best expert. We emphasize that the incorporation of the rule-based model into MoE would not have been possible without generalizing the OCO framework to handle the nonconvex and nonsmooth minFRDE$_k$ loss.

\subsection{Nonstationary Environment}
\input{sections/fig-nonstationary}

We then let the distribution shifts happen three times, at roughly timestep 25,000, 60,000, and 130,000 from Las Vegas to Boston to Singapore to Pittsburgh.

{\bf Convex Loss}.
Fig.~\ref{fig:non-stationary}(a) plots the result using the convex probability loss. We observe that (i) the \squint algorithm rapidly adapts the probability vector during shifts (Fig.~\ref{fig:non-stationary}(a) bottom). In each distribution shift, the algorithm quickly converges to the best expert within that time window. Notably, the \squint algorithm accomplishes this without any prior knowledge of when the distribution shifts occur. (ii) The result of this rapid adaptation is shown in the performance plots (Fig.~\ref{fig:non-stationary}(a) top and middle). The MoE model's performance \emph{tracks the performance of the best expert}. Though the performance of each singular model fluctuates, the MoE's performance is always close to the best singular model.

{\bf Nonconvex Loss}.
When using the nonconvex loss, the probability vector does not converge and it is hard to observe a pattern in the time series of probability vectors (Fig.~\ref{fig:non-stationary}(b) bottom). However, the performance of the MoE model remains robust to distribution shifts, consistently achieving near-optimal performance at all time steps.

%% file: sections/fig-stationary.tex

\begin{figure}[t]
    \begin{minipage}{\columnwidth}
        \begin{tabular}{cc}
            \hspace{-8mm}
            \begin{minipage}{0.5\textwidth}
            \centering
            \includegraphics[width=\columnwidth]{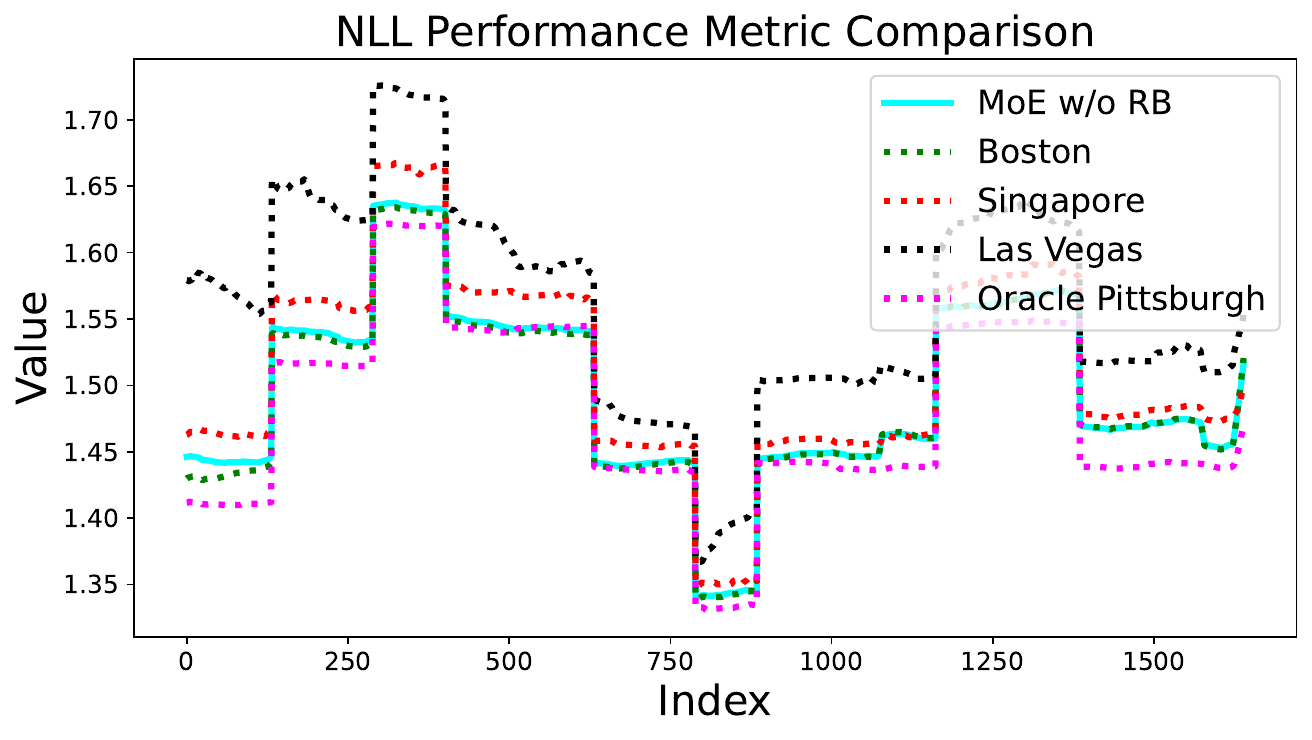}
            \end{minipage}
            &
            \begin{minipage}{0.5\textwidth}
            \centering
            \includegraphics[width=\columnwidth]{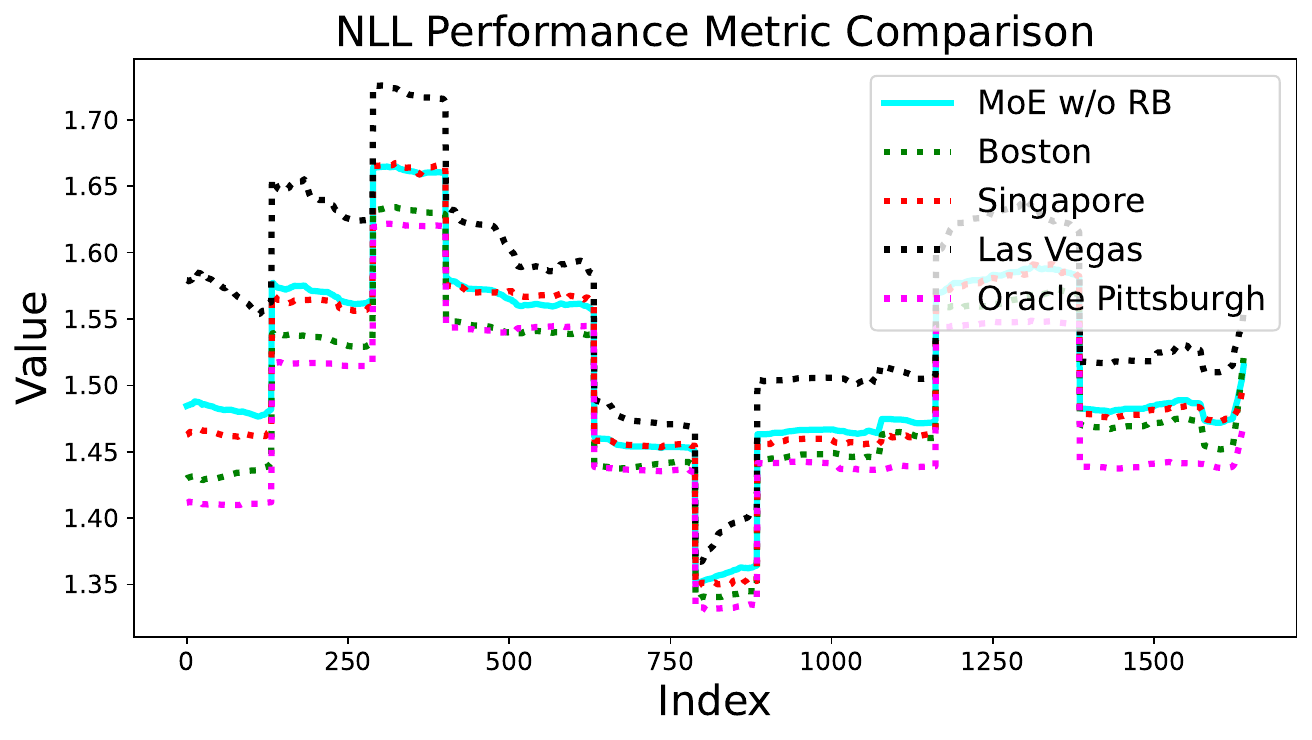}
            \end{minipage}
            \\
            \hspace{-8mm}
            \begin{minipage}{0.5\textwidth}
            \centering
            \includegraphics[width=\columnwidth]{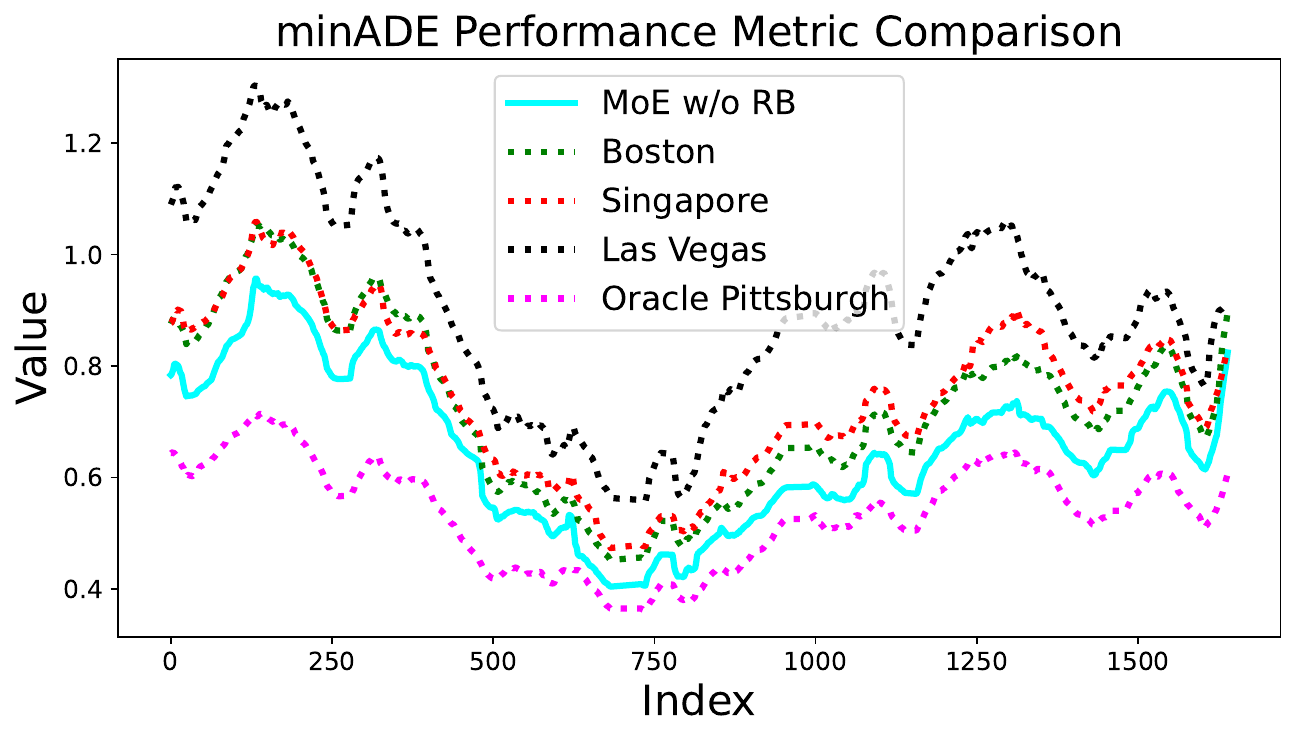}
            \end{minipage}
            &
            \begin{minipage}{0.5\textwidth}
            \centering
            \includegraphics[width=\columnwidth]{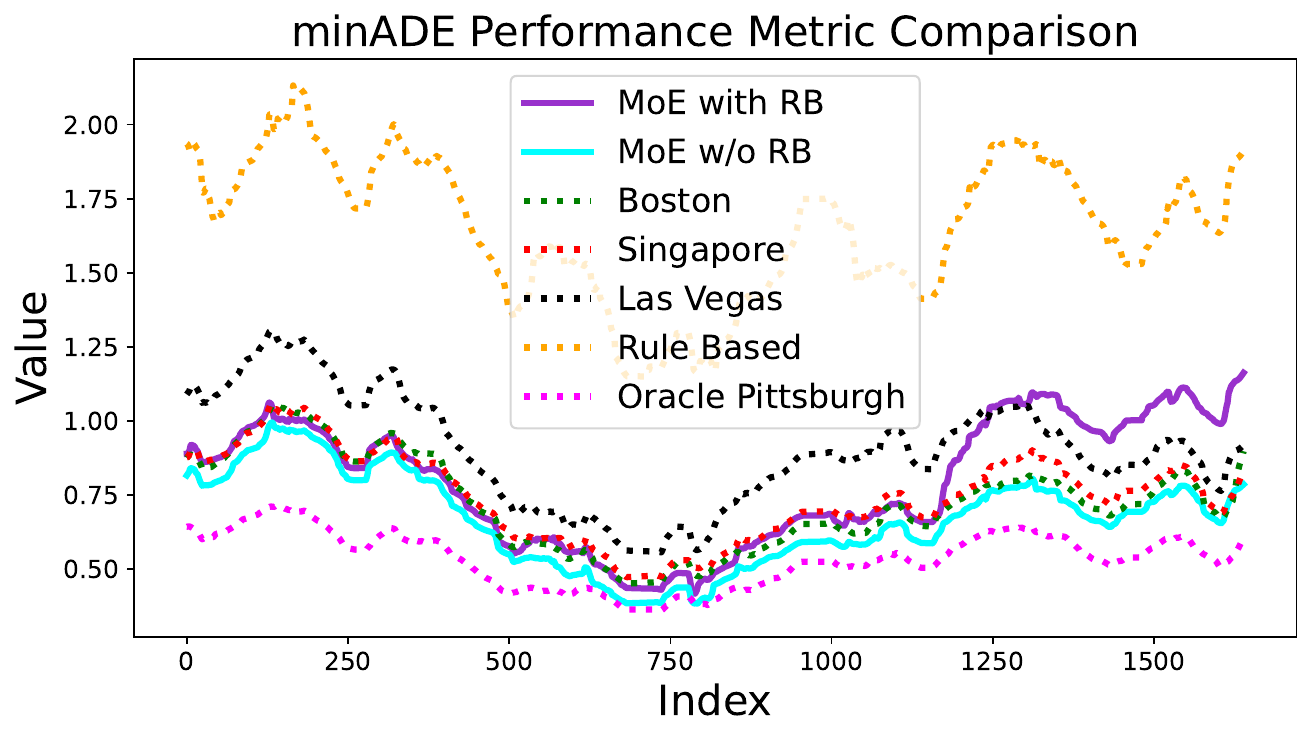}
            \end{minipage}
            \\
            \hspace{-8mm}
            \begin{minipage}{0.5\textwidth}
            \centering
            \includegraphics[width=\columnwidth]{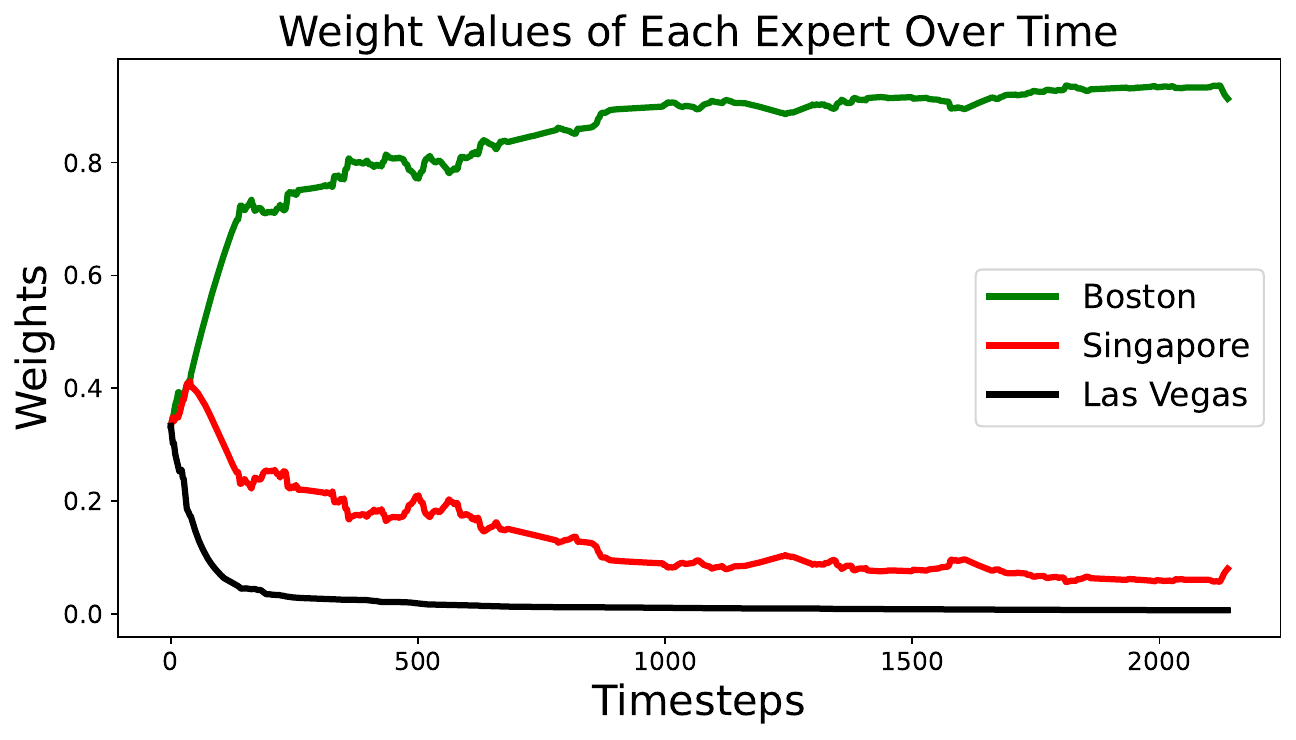} \\
            {(a) Convex probability loss~\eqref{eq:probability-loss}}
            \end{minipage}
            &
            \begin{minipage}{0.5\textwidth}
            \centering
            \includegraphics[width=\columnwidth]{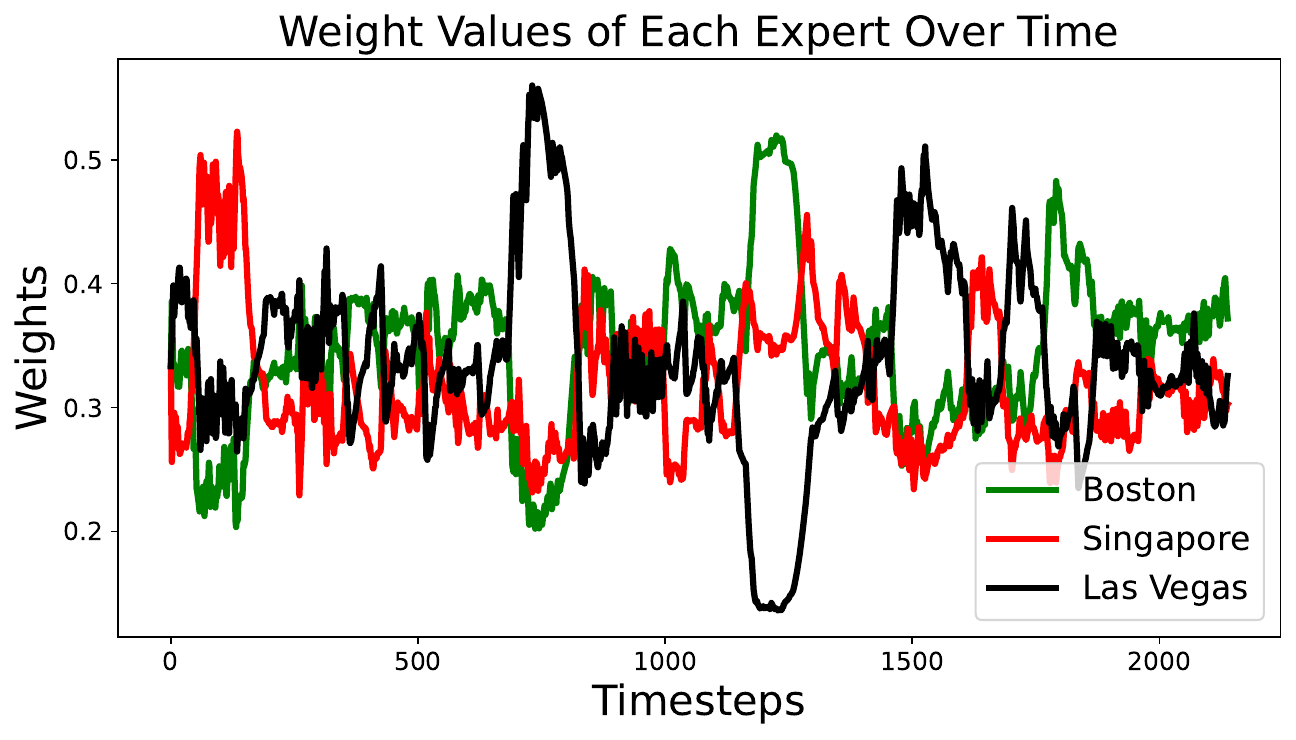} \\
            {(b) Nonconvex minFRDE$_k$ loss~\eqref{eq:minFRDEk}}
            \end{minipage} 
        \end{tabular}
    \end{minipage}
    \caption{Performance of MoE in a stationary distribution shift (from the Pittsburgh environment) using (a) convex loss and (b) nonconvex loss. From top to bottom: NLL performance of the MoE compared with the singular models; minADE performance of the MoE compared with the singular models; evolution of the probability vector. minFDE performance is shown in Appendix~\ref{app:experiments}.
    \label{fig:stationary}}
    \vspace{2mm}
\end{figure}

%% file: sections/fig-lyft.tex

\begin{figure}[t]
    \begin{minipage}{\columnwidth}
        \begin{tabular}{cc}
            \hspace{-8mm}
            \begin{minipage}{0.5\textwidth}
            \centering
            \includegraphics[width=\columnwidth]{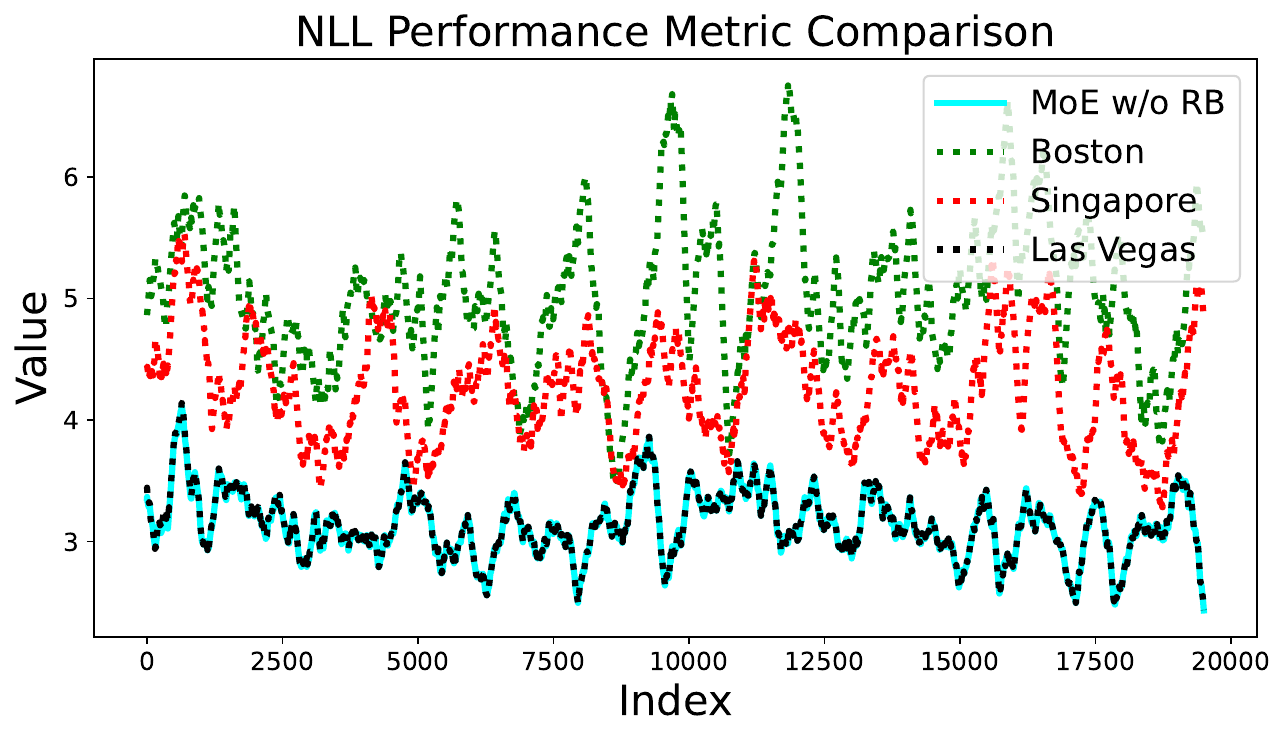}
            \end{minipage}
            &
            \begin{minipage}{0.5\textwidth}
            \centering
            \includegraphics[width=\columnwidth]{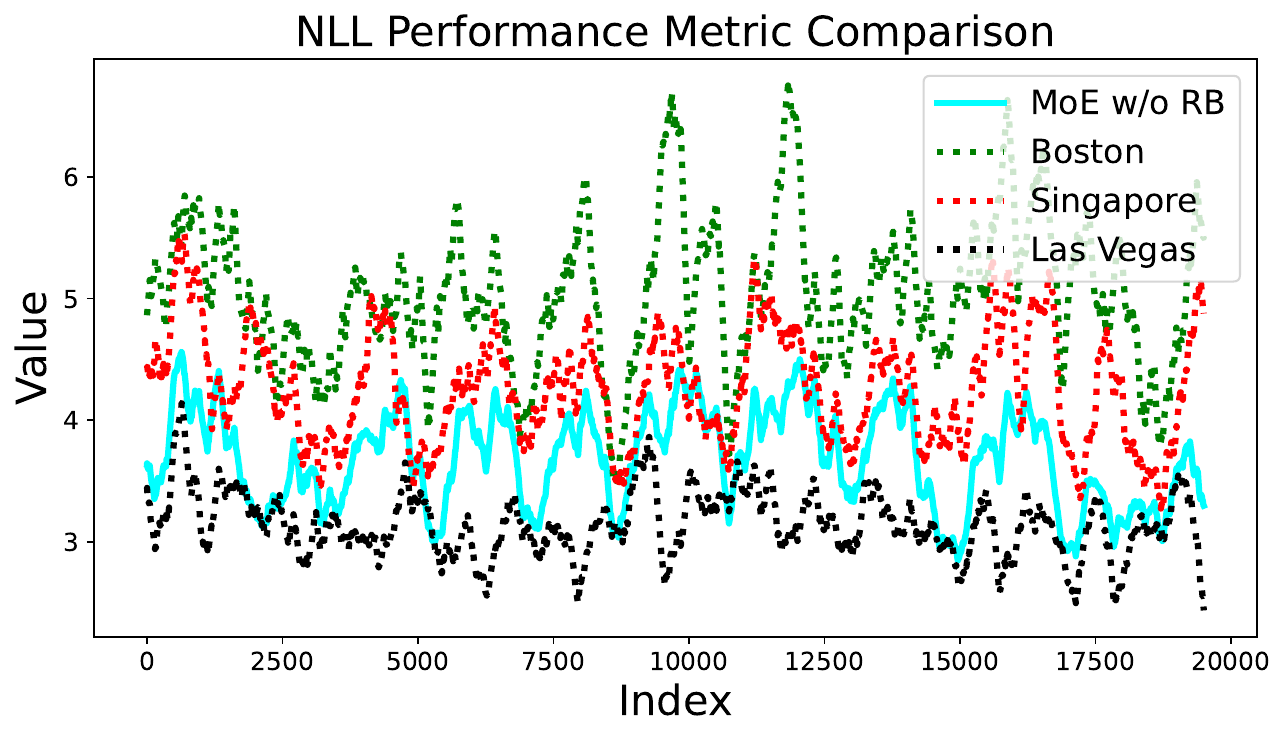}
            \end{minipage}
            \\
            \hspace{-8mm}
            \begin{minipage}{0.5\textwidth}
            \centering
            \includegraphics[width=\columnwidth]{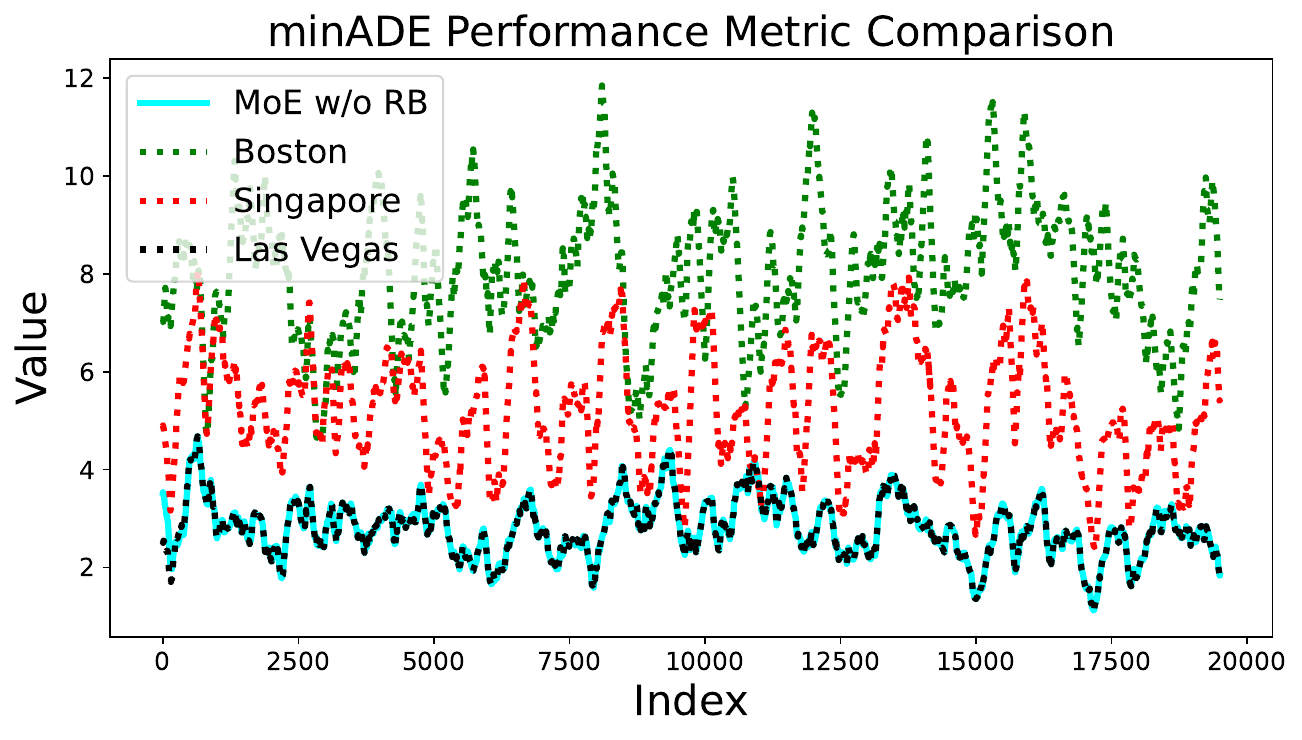}
            \end{minipage}
            &
            \begin{minipage}{0.5\textwidth}
            \centering
            \includegraphics[width=\columnwidth]{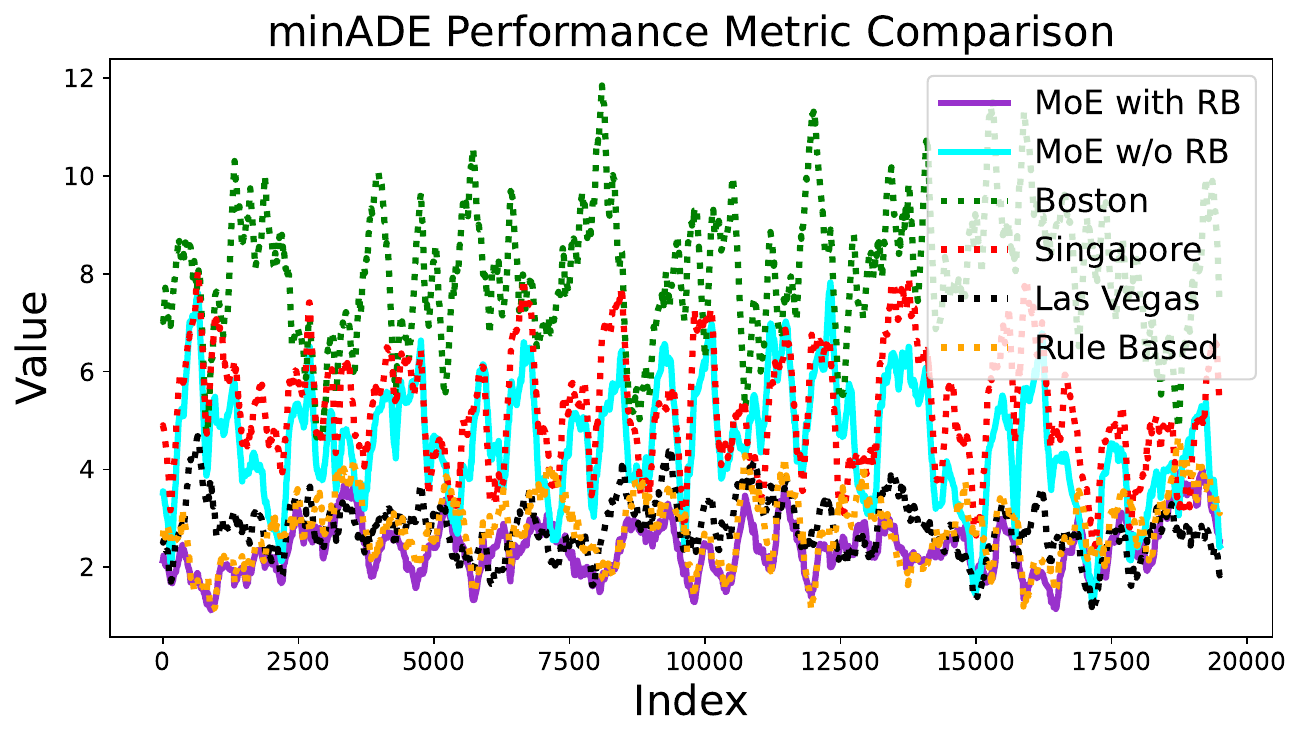}
            \end{minipage}
            \\
            \hspace{-8mm}
            \begin{minipage}{0.5\textwidth}
            \centering
            \includegraphics[width=\columnwidth]{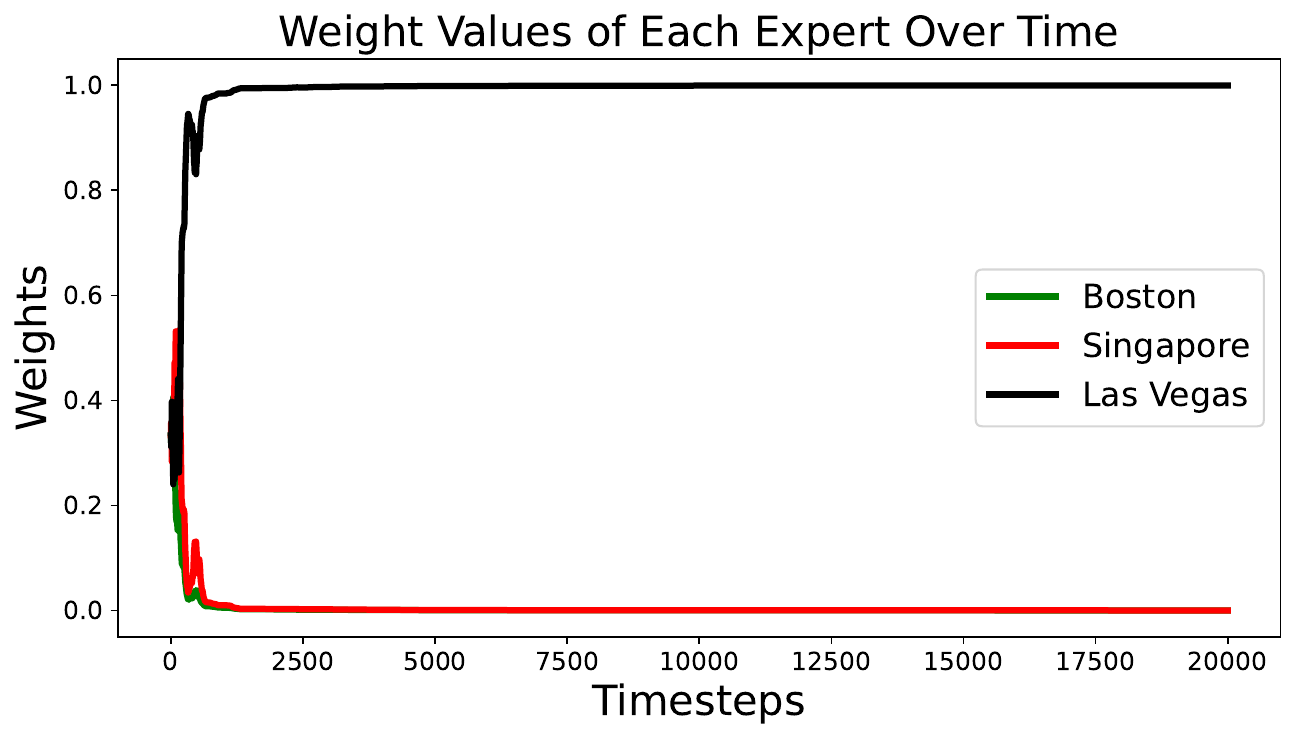} \\
            {(a) Convex probability loss~\eqref{eq:probability-loss}}
            \end{minipage}
            &
            \begin{minipage}{0.5\textwidth}
            \centering
            \includegraphics[width=\columnwidth]{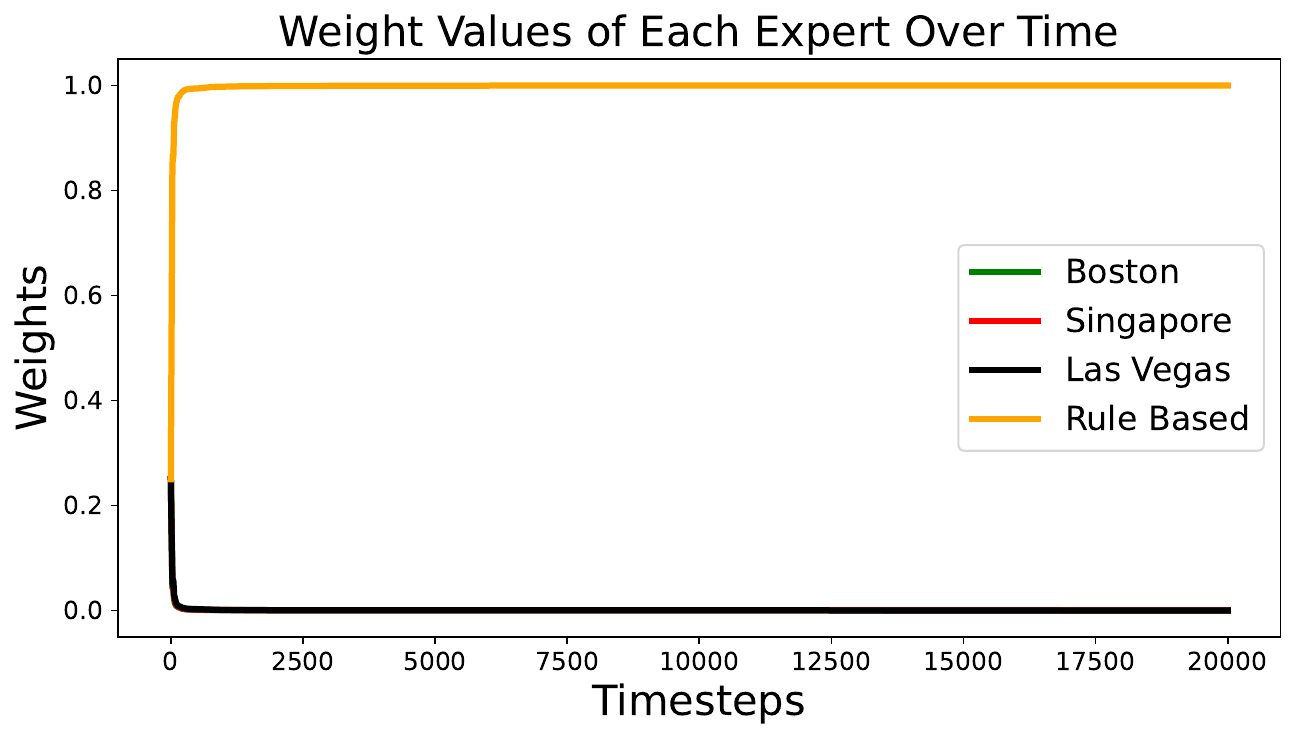} \\
            {(b) Nonconvex minFRDE$_k$ loss~\eqref{eq:minFRDEk}}
            \end{minipage} 
        \end{tabular}
    \end{minipage}
    \caption{Performance of MoE in a stationary distribution shift (from the \lyft dataset) using (a) convex loss and (b) nonconvex loss. From top to bottom: NLL performance of the MoE compared with the singular models; minADE performance of the MoE compared with the singular models; evolution of the probability vector.}
    \label{fig:lyft}
    \vspace{2mm}
\end{figure}

%% file: sections/fig-nonstationary.tex

\begin{figure}[t]
    \begin{minipage}{\columnwidth}
        \begin{tabular}{cc}
            \hspace{-8mm}
            \begin{minipage}{0.5\textwidth}
            \centering
            \includegraphics[width=\columnwidth]{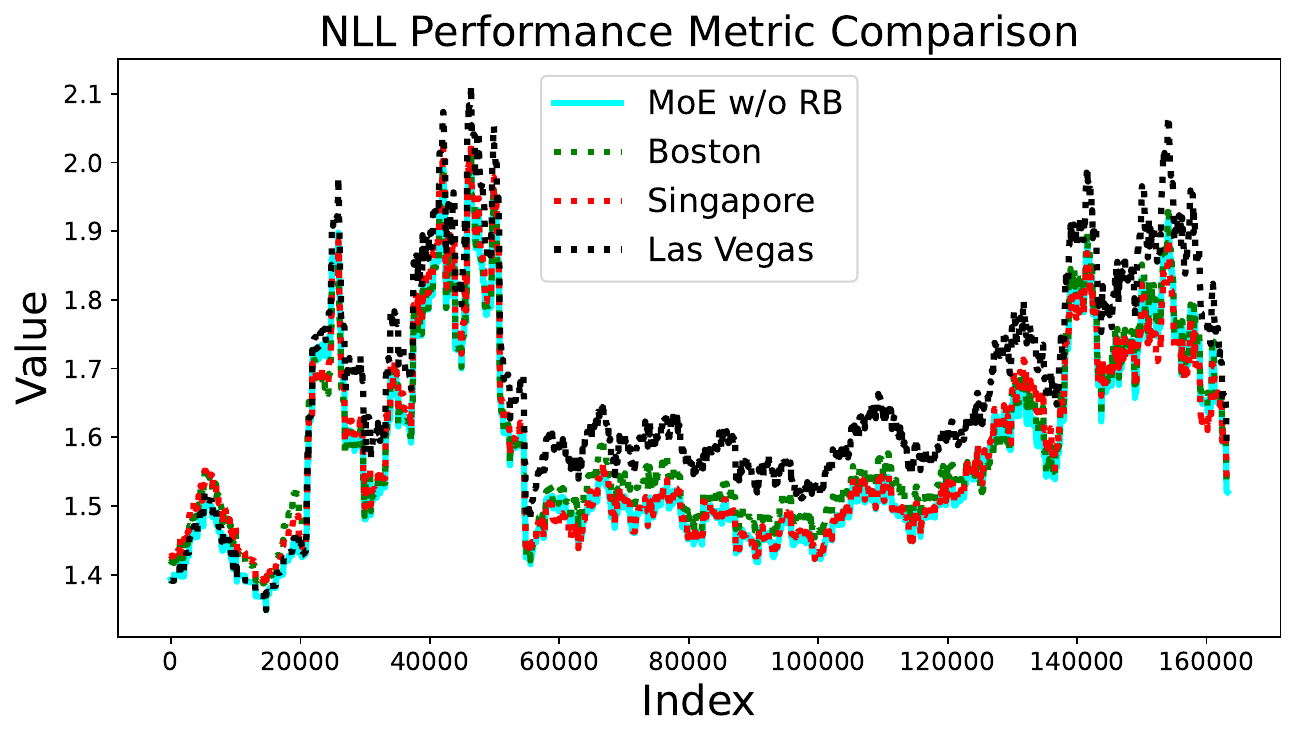}
            \end{minipage}
            &
            \begin{minipage}{0.5\textwidth}
            \centering
            \includegraphics[width=\columnwidth]{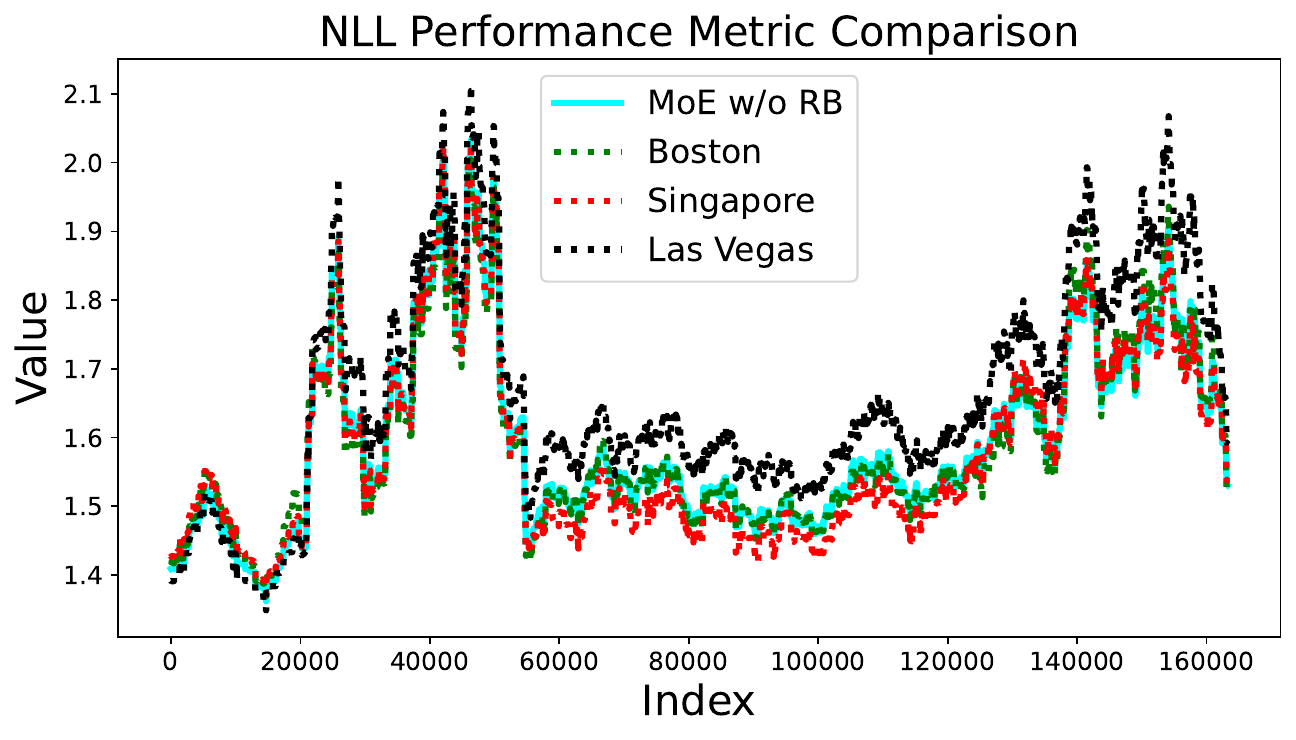}
            \end{minipage}
            \\
            \hspace{-8mm}
            \begin{minipage}{0.5\textwidth}
            \centering
            \includegraphics[width=\columnwidth]{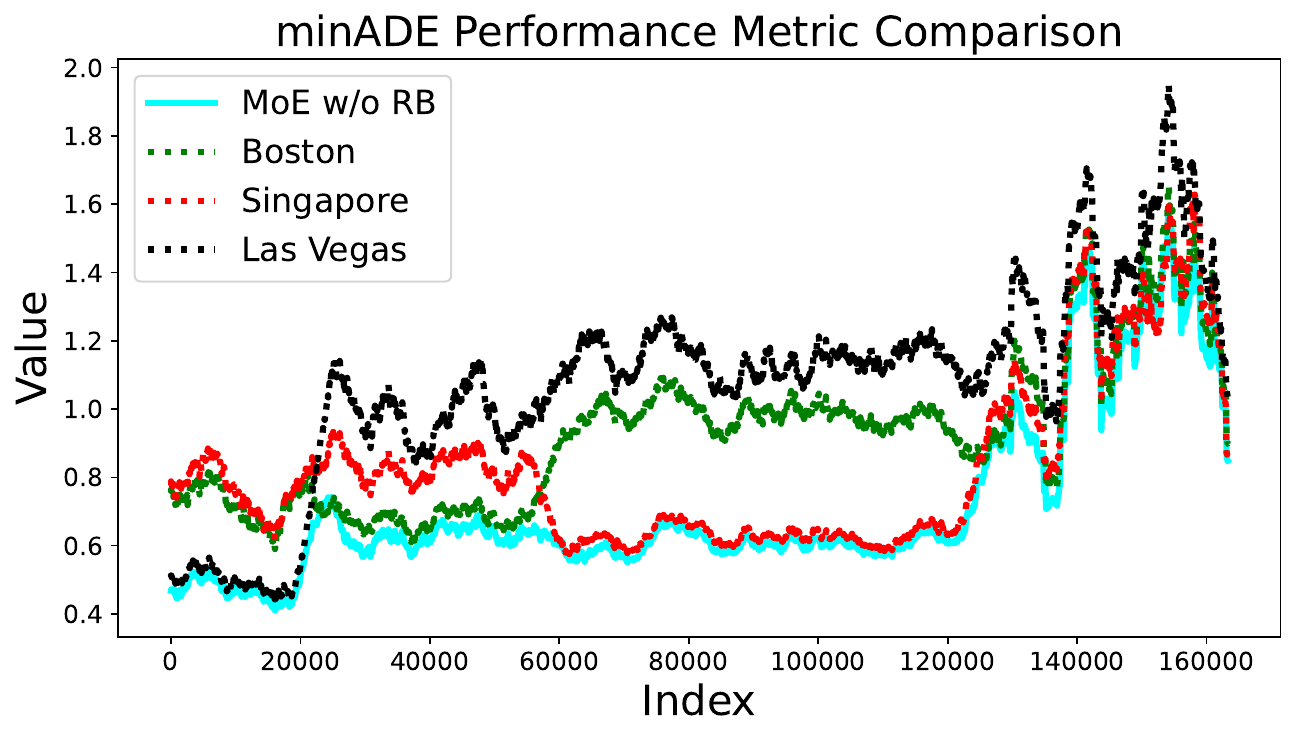}
            \end{minipage}
            &
            \begin{minipage}{0.5\textwidth}
            \centering
            \includegraphics[width=\columnwidth]{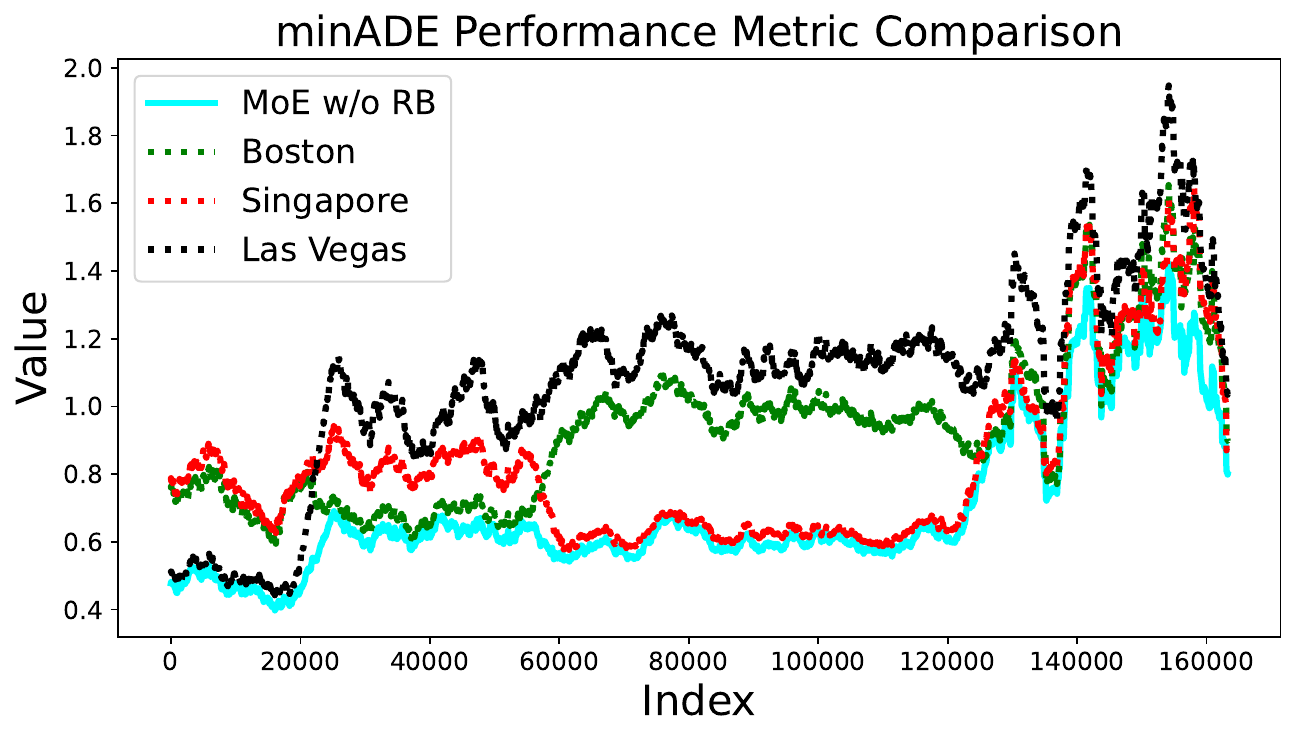}
            \end{minipage}
            \\
            \hspace{-8mm}
            \begin{minipage}{0.5\textwidth}
            \centering
            \includegraphics[width=\columnwidth]{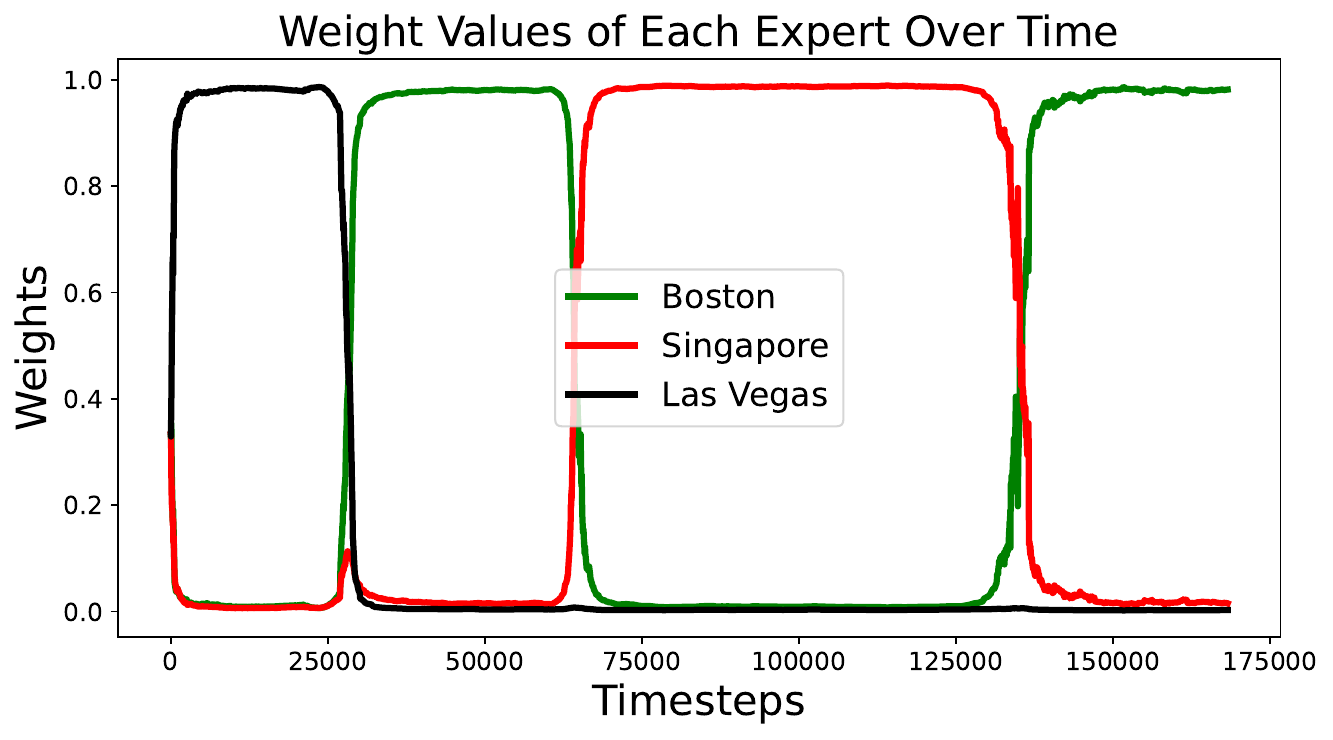} \\
            {(a) Convex probability loss~\eqref{eq:probability-loss}}
            \end{minipage}
            &
            \begin{minipage}{0.5\textwidth}
            \centering
            \includegraphics[width=\columnwidth]{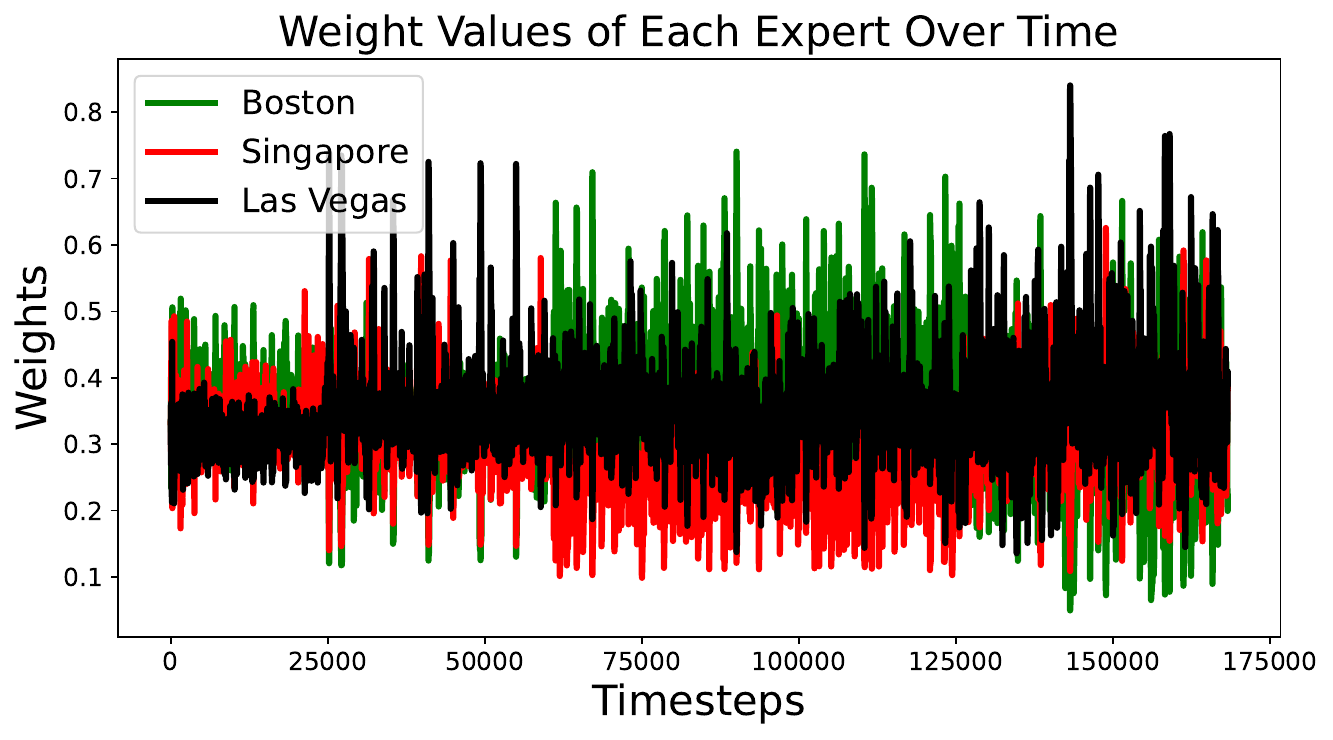} \\
            {(b) Nonconvex minFRDE$_k$ loss~\eqref{eq:minFRDEk}}
            \end{minipage} 
        \end{tabular}
    \end{minipage}
    \caption{Performance of MoE in nonstationary distribution shifts using (a) convex loss and (b) nonconvex loss. From top to bottom: NLL performance of the MoE compared with the singular models; minADE performance of the MoE compared with the singular models; evolution of the probability vector. minFDE performance is shown in Appendix~\ref{app:experiments}. 
    \label{fig:non-stationary}}
    \vspace{2mm}
\end{figure}

%% file: sections/conclusion.tex
\section{Conclusion}
\label{sec:conclusion}

We draw from Online Convex Optimization theory and introduce a light-weight and model-agnostic framework to aggregate multiple trajectory predictors online. Our framework leverages the \squint algorithm originally designed for hedging and extends it to the case of nonconvex and nonstationary environments. Results on the \nuscenes and the \lyft datasets show that our framework effectively combines multiple singular models to form a stronger mixture model.

{\bf Limitation}.
(i) Aggregating multiple models comes at the cost of running several models online. While this overhead can be mitigated through GPU parallelization, future work should explore using smaller neural networks as individual experts to reduce runtime costs. (ii) Due to the scarcity of data splits, we aggregated only three and four models. Future work should investigate the aggregation of tens or even hundreds of models for trajectory prediction or other applications.

%% file: sections/appendix.tex

\appendix

\subsection{Exponentiated Gradient and Comparison with \squint}
\label{app:eg-squint}

The \emph{exponentiated gradient} (EG) algorithm~\cite{orabona19book-modern} is presented in Algorithm~\ref{alg:EG} (See~\cite{uwashington_lecture_10_2012} for a brief introduction).

\vspace{-3mm}
\input{sections/alg-exponentiated-gradient}
\vspace{-3mm}

\begin{figure}[h]
    \begin{minipage}{\columnwidth}
        \begin{tabular}{cc}
            \hspace{-4mm}
            \begin{minipage}{0.5\textwidth}
            \centering
            \includegraphics[width=\columnwidth]{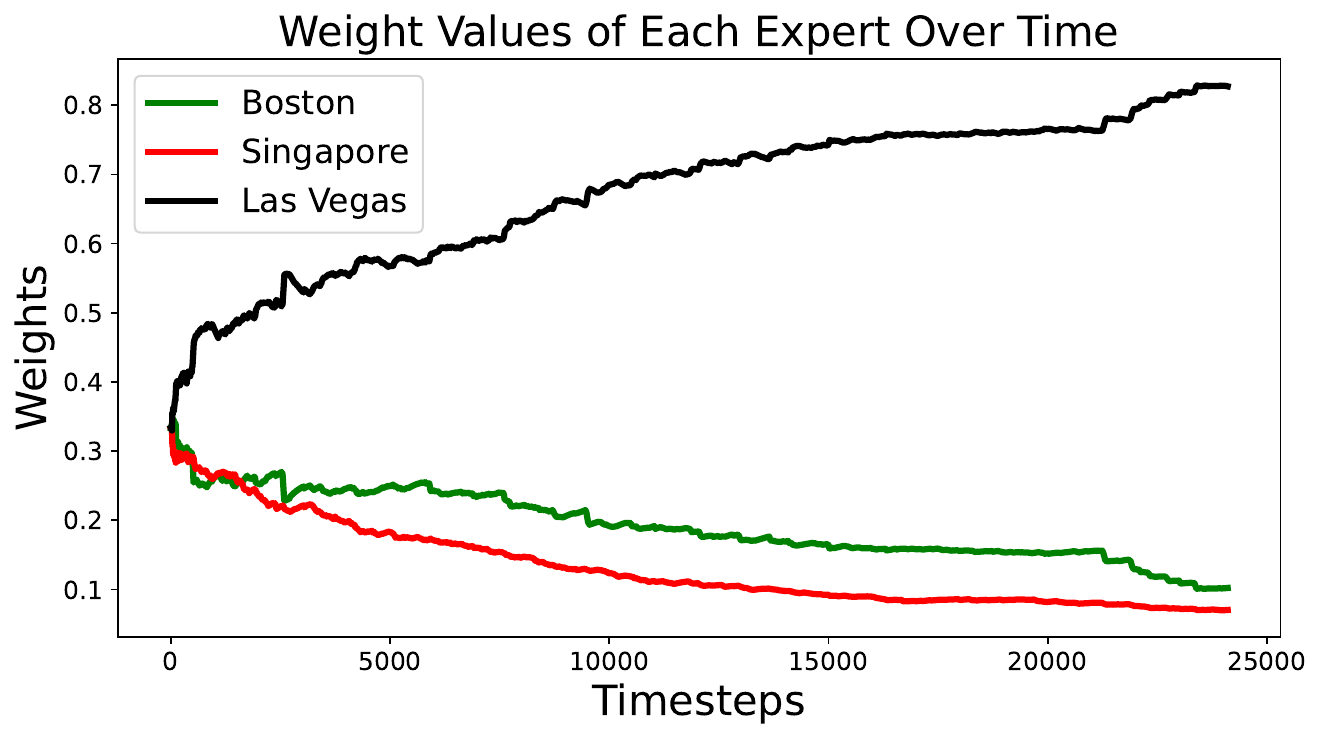}
            {(a) Exponentiated Gradient}
            \end{minipage}
            &
            \hspace{-4mm}\begin{minipage}{0.5\textwidth}
            \centering
            \includegraphics[width=\columnwidth]{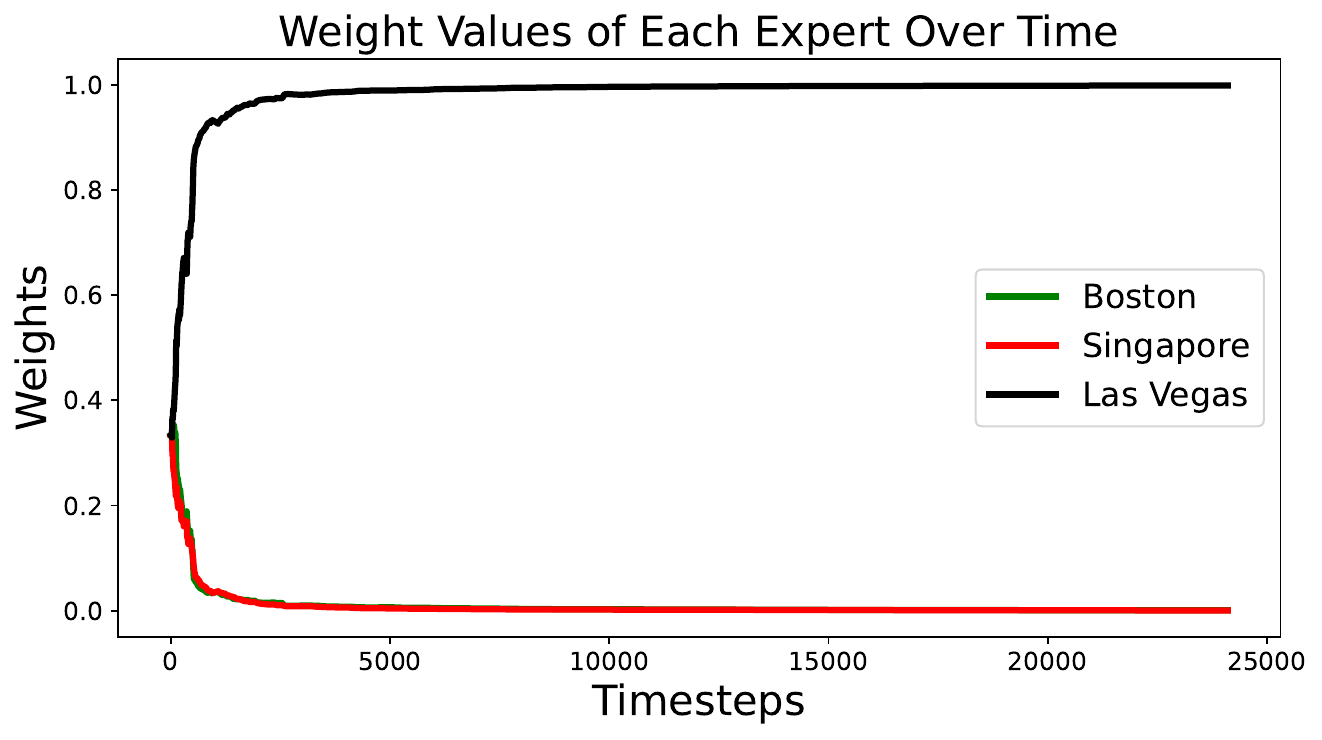}
            {(b) \squint}
            \end{minipage}
            \\
        \end{tabular}
    \end{minipage}
    \caption{\squint vs EG on the Las Vegas dataset.
    \label{fig:squintvsEG}}
\end{figure}

We evaluated both the EG algorithm and \squint on the Las Vegas dataset, where we expect the algorithms to converge on choosing the Las Vegas model as the best expert. Fig.~\ref{fig:squintvsEG} shows the evolution of the probability vector $\alpha_t$ over time for (a) EG and (b) \squint. Notably, \squint converges approximately 25 times faster than EG, demonstrating a significant improvement in adaptation speed.

\vspace{-2mm}

\subsection{Experimental Details}
For the stationary environment, we evaluate our models using the \texttt{nuplan_mini-pittsburgh} dataset from \nuscenes. To test performance under nonstationary conditions, we concatenate multiple datasets together in sequence: \texttt{nuplan_mini-las_vegas}, \texttt{nuplan_val-boston}, \texttt{nuplan_val-singapore}, \texttt{nuplan_val-pittsburgh}.

We use $\beta=10$ as the temperature scaling factor in $\mathrm{softmin}$.

We set the discount factor $\lambda=0.9999$ in the nonstationary setup.

We use $k=10$ in choosing the top-$k$ modes.

\subsection{Further Experimental Results}
\label{app:experiments}

\begin{figure}[h]
    \begin{minipage}{\columnwidth}
        \begin{tabular}{cc}
            \hspace{-8mm}
            \begin{minipage}{0.5\textwidth}
            \centering
            \includegraphics[width=\columnwidth]{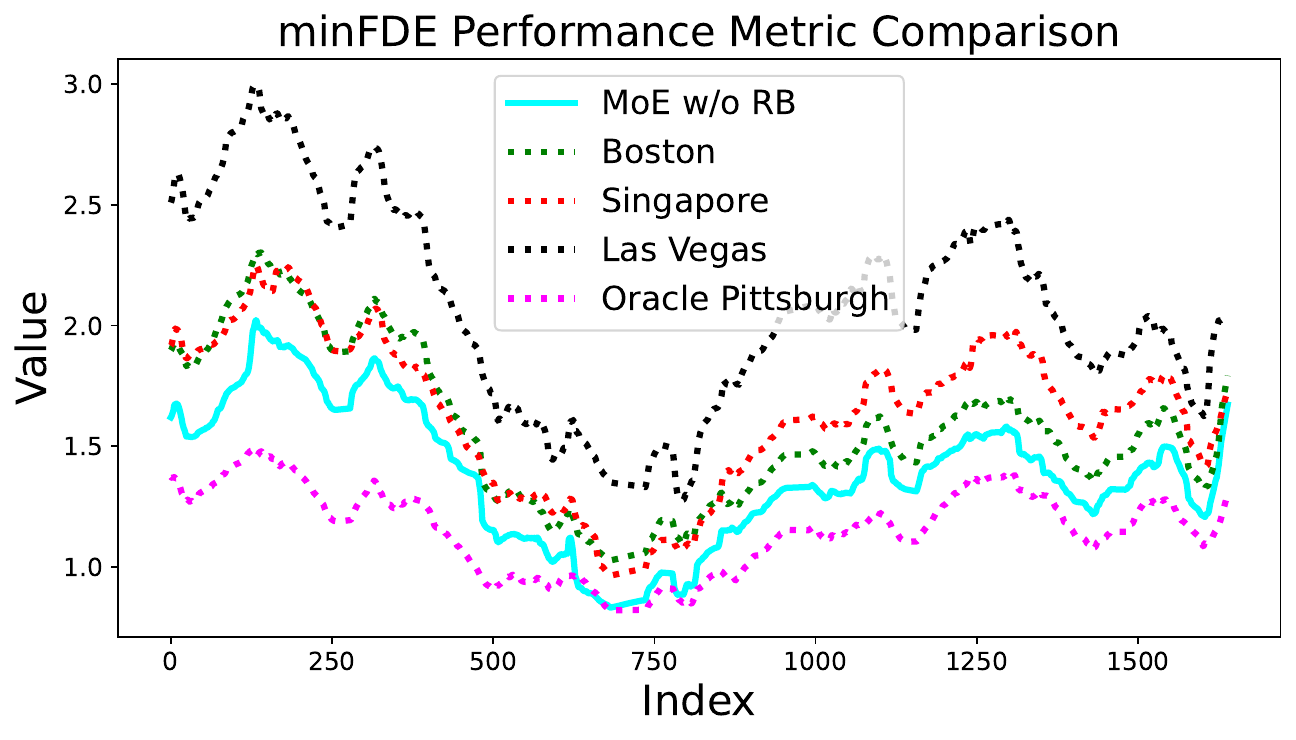}
            \end{minipage}
            &
            \begin{minipage}{0.5\textwidth}
            \centering
            \includegraphics[width=\columnwidth]{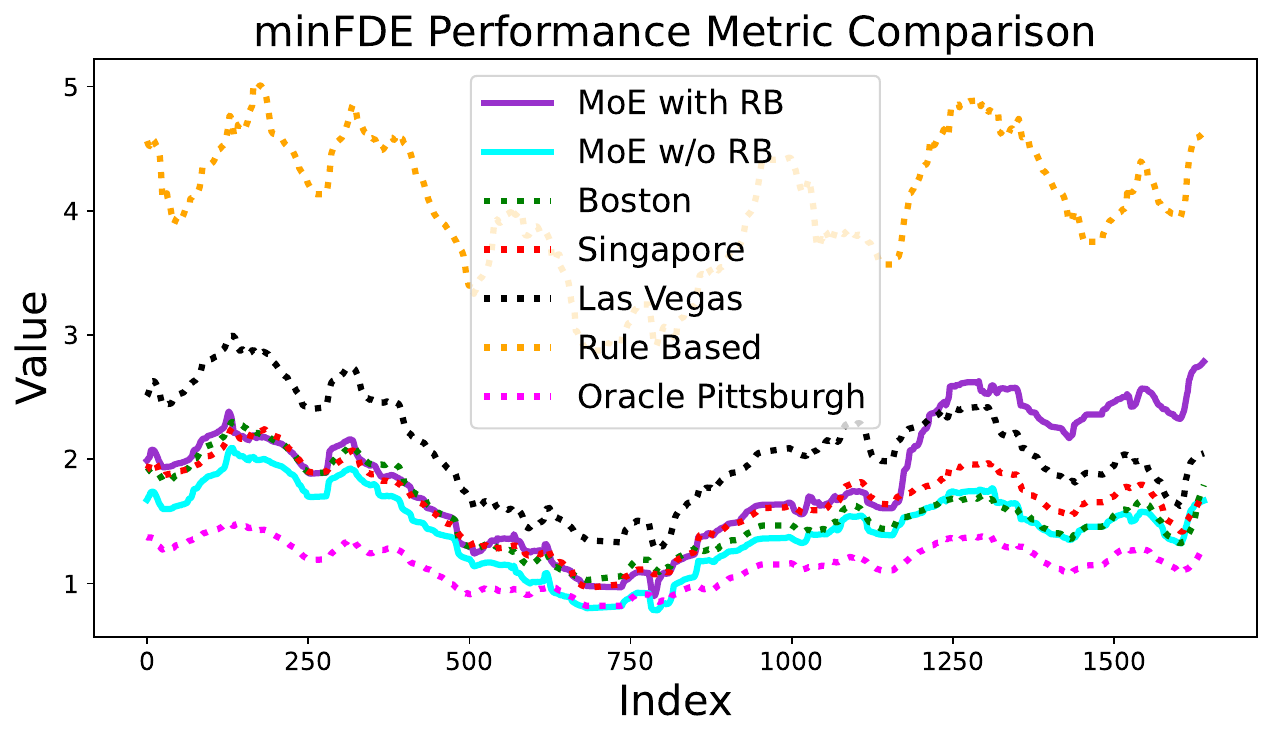}
            \end{minipage}
            \\
            \hspace{-8mm}
            \begin{minipage}{0.5\textwidth}
            \centering
            \vspace{26.5mm}
            {(a) Convex probability loss~\eqref{eq:probability-loss}}
            \end{minipage}
            &
            \begin{minipage}{0.5\textwidth}
            \centering
            \includegraphics[width=\columnwidth]{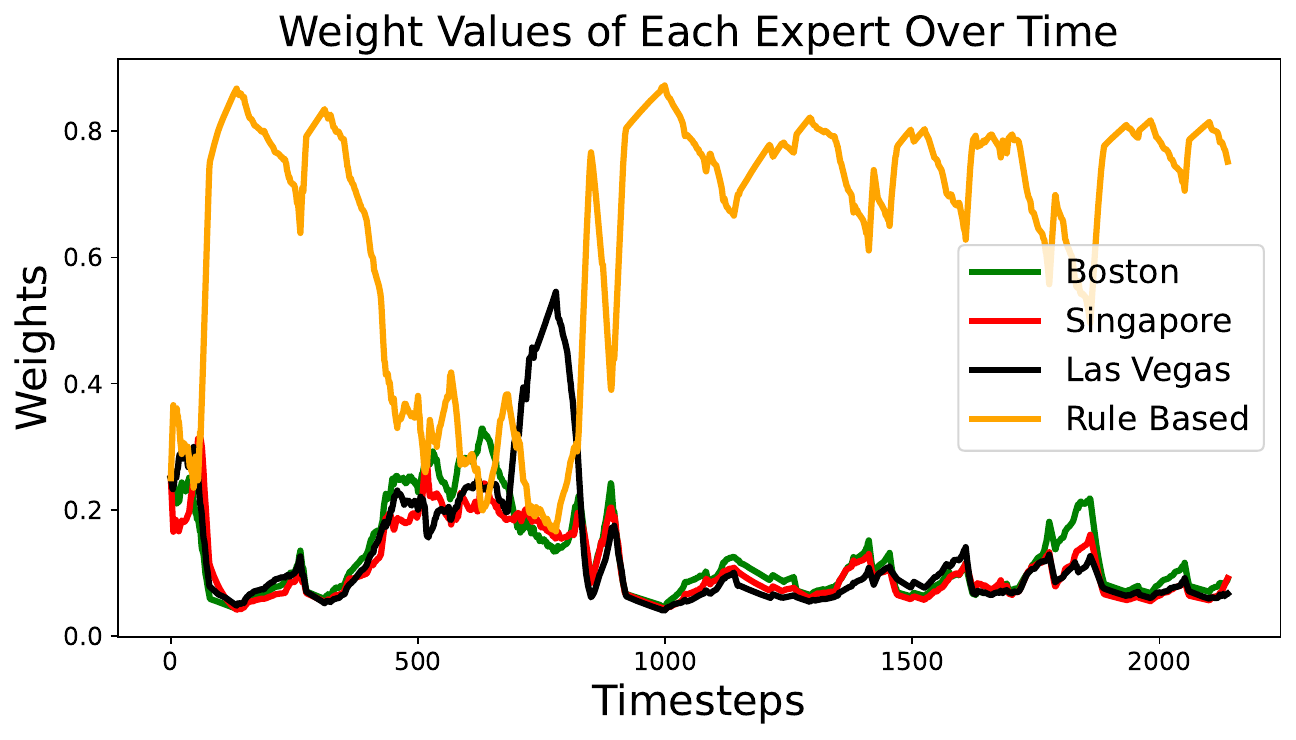} 
            {(b) Nonconvex minFRDE$_k$ loss~\eqref{eq:minFRDEk} with RB}
            \end{minipage}
            \\
        \end{tabular}
    \end{minipage}
    \caption{More results on the performance of MoE in a stationary setting using (a) probability loss and (b) minFRDE loss. (a) shows the minFDE performance metric. (b) top shows the minFDE performance metric and (b) bottom shows the evolution of the probability vector when aggregating three learned predictors and one rule-based predictor. 
    \label{fig:stationary-extra}}
    \vspace{2mm}
\end{figure}

Fig.~\ref{fig:stationary-extra} plots extra results in the stationary setup.

\begin{figure}[h]
    \begin{minipage}{\columnwidth}
        \begin{tabular}{cc}
            \hspace{-8mm}
            \begin{minipage}{0.5\textwidth}
            \centering
            \includegraphics[width=\columnwidth]{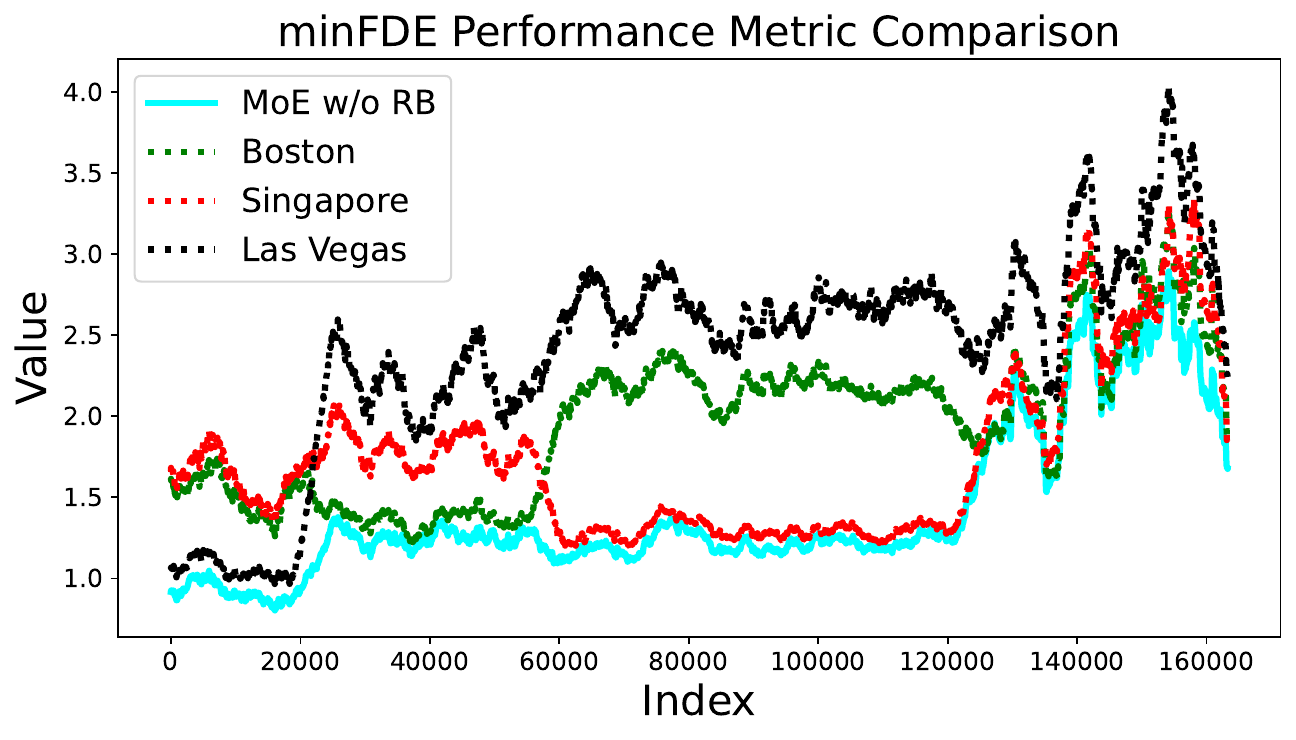}
            {(a) Convex probability loss~\eqref{eq:probability-loss}}
            \end{minipage}
            &
            \begin{minipage}{0.5\textwidth}
            \centering
            \includegraphics[width=\columnwidth]{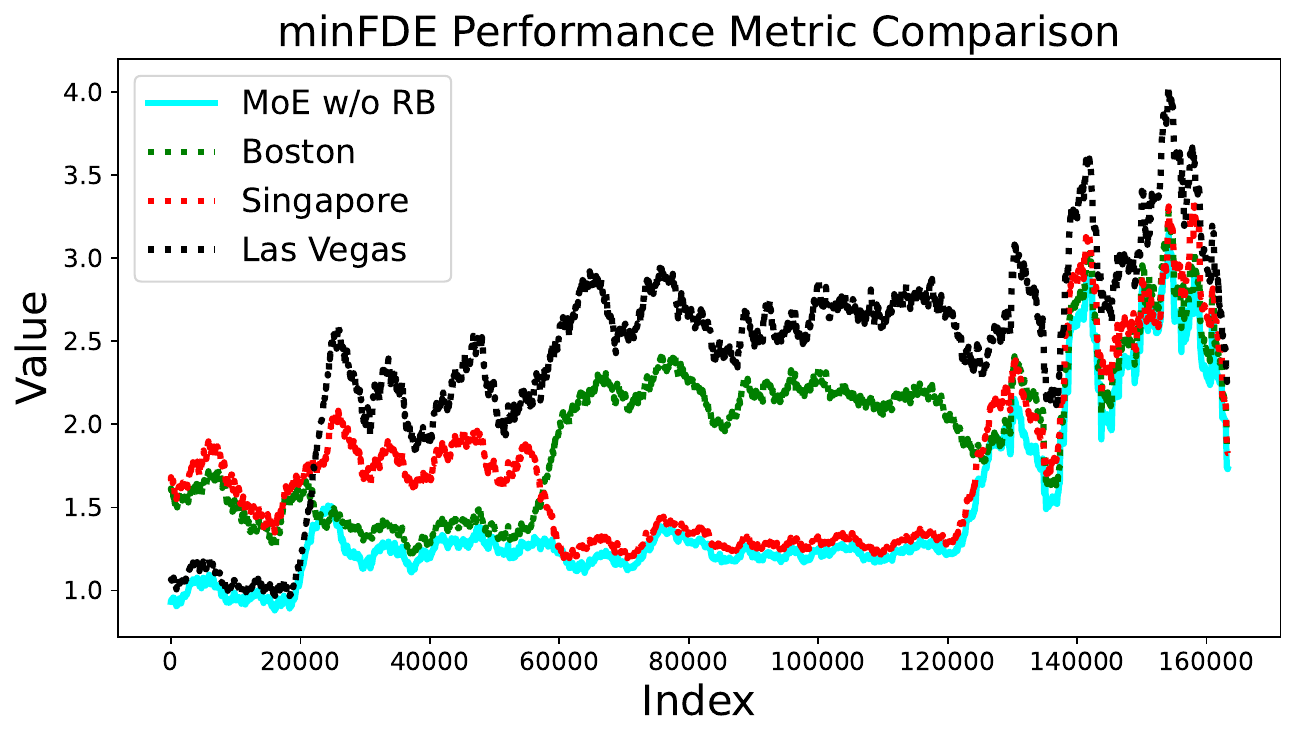}
            {(b) Nonconvex minFRDE$_k$ loss~\eqref{eq:minFRDEk}}
            \end{minipage}
            \\
        \end{tabular}
    \end{minipage}
    \caption{minFDE performance metric in a nonstationary setting using (a) convex probability loss and (b) nonconvex minFRDE loss. 
    \label{fig:nonstationary-extra}}
    \vspace{2mm}
\end{figure}

Fig.~\ref{fig:nonstationary-extra} plots extra results in the nonstationary setup.

At first glance, the behavior shown in Fig.~\ref{fig:stationary-extra} may seem counterintuitive since the probability vector favors the rule-based expert despite its poor performance. However, this preference stems from the rule-based expert's underlying design: it first predicts the agent's desired lane and then plans to maintain that position. While this strategy is typically effective, prediction errors jump up drastically when the initial lane choice is incorrect. This is then heavily emphasized by the sliding window averaging we employ, making the rule-based expert's overall performance appear significantly worse than other experts, even though it performs well in most cases where the lane prediction is accurate.

\subsection{Generality of the Framework}
\label{app:generality}

In this section, we elaborate on Remark~\ref{remark:general} and show that our online aggregation framework presented in Section~\ref{sec:online-agg} can handle cases where the pretrained predictors output arbitrary distributions (\ie not necessarily GMMs), as long as they can be sampled from.

{\bf Problem Setup} At time $t$, we are given $N$ expert trajectory predictors, where each predictor predicts an arbitrary distribution of trajectories. We assume that this arbitrary distribution can be sampled from, and, in particular, the $i$-th expert generates $M$ trajectory samples
\bea 
\calT^i_t = \{ x^i_{1,t}, x^i_{2,t},\dots, x^i_{M,t} \}, \quad i=1,\dots,N,
\eea
where the superscript $i$ denotes the expert index. Similar to Section~\ref{sec:online-agg}, we want to choose a probability vector $\alpha_t \in \Delta_N$, where each $\alpha_{i,t}$ represents the weight assigned to the $i$-th trajectory predictor. 

{\bf Loss Function for OCO} To evaluate the performance of our MoE, a natural strategy is to compute a suitable loss between the individual trajectory samples $\calT^i_t,i=1,\dots,N$ and the groundtruth trajectory $x_t$, and then seek to minimize the weighted sum of the individual losses. Formally, at time $t$, we design the loss function of $\alpha_t$ as
\bea \label{eq:loss-samples}
\ell_t(\alpha_t) = \sum_{i=1}^N \alpha_{i,t} \ell_{t}^i (\calT^i_t, x_t),
\eea 
where the individual loss of each expert, $\ell_t^i(\cdot)$, is any meaningful loss function that computes the performance of the $i$-th expert. The simplest choice would be
\bea 
\ell^i_t (\calT^i_t, x_t) = \frac{1}{M} \sum_{j=1}^M \Vert x^i_{j,t} - x_t \Vert^2, 
\eea 
where the mean squared error is computed. Another choice, if the designer is only concerned with the ``good'' trajectory samples, is to pick the best $k$ samples and compute their mean squared error, \ie 
\bea 
\ell^i_t (\calT^i_t, x_t) = \frac{1}{k} \sum_{j=1}^k \mathrm{topk} \left\{ \Vert x^i_{j,t} - x_t \Vert^2 \right\}_{j=1}^M, 
\eea
where the function $\mathrm{topk}(\cdot)$ selects the $k$ minimum scalars. 

{\bf \squint}. A crucial observation is that the loss function $\ell_t(\alpha_t)$ in~\eqref{eq:loss-samples} remains \emph{a linear function} of $\alpha_t$. Therefore, \squint presented in Algorithm~\ref{alg:squint} can still be applied without any modifications.

{\bf Mixture of Experts from Importance Sampling}. How do we form the Mixture-of-Experts (MoE) model in this case? The answer is to use importance sampling. Specically, the MoE model consists of a new set of $M$ trajectory samples drawn from the mixture of distributions (each generated by one expert), with the mixing weights being $\alpha_t$. One way to achieve this is to draw $\lfloor M \alpha_{i,t} \rfloor$ random samples from the $i$-th expert. Intuitively, if the weight of the first expert $\alpha_{1,t}$ decreases to near zero (due to the loss $\ell^1_t(\calT^1_t, x_t)$ being very large), then the MoE samples will effectively exclude the first expert. 

{\bf Comparison with GMM}. After the discussion above, we see that Online Convex Optimization (OCO) can be applied to the online aggregation of arbitrary distributions, as long as they can be sampled from. Moreover, the algorithm remains unchanged, with the only difference being in the computation of the loss functions. However, we emphasize that a key drawback of the loss function~\eqref{eq:loss-samples} is its reliance on sampling from the distributions, which may require a large number of samples $M$ to ensure fast convergence.

%% file: sections/alg-exponentiated-gradient.tex
\setlength{\textfloatsep}{0pt}
\begin{algorithm}[hbt!]
\SetKwInput{KwInput}{Input}
\SetKwInput{KwInit}{Initialize}
\caption{Exponentiated Gradient}\label{alg:EG}
\KwInput{A prior distribution of the weights $\alpha_1$}
\For{$t=1,2,\dots, T$}{
    Compute $\ell_t$ as in \eqref{eq:probability-loss} and define the gradient $\tilde{g}_t = \nabla_{\alpha_t}(\ell_t) \in \Real{N}$ 
    
    Set $G_t = \max (\max_{i \in [N]}(|\tilde{g}_{t,i}|), G_{t-1})$ ($\tilde{g}_{t,i}$ denotes the $i$-th element of $\tilde{g}_t$) \label{line:Gt-EG}
    
    Define the clipped gradient $g_t = \frac{(\tilde{g}_t / G_t) + 1}{2} \in \Real{N}$ (element-wise division) \label{line:clip-gradient-EG}
    
    Update 
    \bea\label{eq:eg-update-alpha}
    w_{t+1,i} = \exp \parentheses{-\frac{\sum_{\tau=1}^{t-1}g_\tau(i) \sqrt{\ln N}}{\sqrt{t}}}
    \eea
    
    Update $\alpha_{t+1} = \frac{w}{\sum_{i=1}^N w_i}$ \label{line:normweights}
}
\end{algorithm}